\documentclass[sigconf]{acmart}
\usepackage{amsmath,amsfonts,amssymb}
\usepackage{hyperref}
\hypersetup{
    colorlinks=true,
    linkcolor=blue,
    filecolor=magenta,      
    urlcolor=cyan,
}
\usepackage{subcaption}
\usepackage{float}
\usepackage{graphics}
\usepackage{soul}
\usepackage{epstopdf}
\usepackage{multirow}
\usepackage{booktabs}
\usepackage{amsfonts}
\usepackage{booktabs}
\usepackage{algorithm}
\usepackage{algorithmic}
\newcommand{\ModelName}{AGE} 
\newcommand\blfootnote[1]{%
  \begingroup
  \renewcommand\thefootnote{}\footnote{#1}%
  \addtocounter{footnote}{-1}%
  \endgroup
}

\AtBeginDocument{%
  \providecommand\BibTeX{{%
    \normalfont B\kern-0.5em{\scshape i\kern-0.25em b}\kern-0.8em\TeX}}}

\copyrightyear{2020}
\acmYear{2020}
\setcopyright{acmlicensed}\acmConference[KDD '20]{Proceedings of the 26th ACM SIGKDD Conference on Knowledge Discovery and Data Mining}{August 23--27, 2020}{Virtual Event, CA, USA}
\acmBooktitle{Proceedings of the 26th ACM SIGKDD Conference on Knowledge Discovery and Data Mining (KDD '20), August 23--27, 2020, Virtual Event, CA, USA}
\acmPrice{15.00}
\acmDOI{10.1145/3394486.3403140}
\acmISBN{978-1-4503-7998-4/20/08}

\acmSubmissionID{959}

\begin{document}
\fancyhead{}
\title{Adaptive Graph Encoder for Attributed Graph Embedding}

\author{Ganqu Cui$^{1,2,3}$, Jie Zhou$^{1,2,3}$, Cheng Yang$^{4*}$, Zhiyuan Liu$^{1,2,3*}$}

\affiliation{%
  \institution{$^1$Department of Computer Science and Technology, Tsinghua University}
  \institution{$^2$Institute for Artificial Intelligence, Tsinghua University}
  \institution{$^3$Beijing National Research Center for
Information Science and Technology}
  \institution{$^4$School of Computer Science, Beijing University of Posts and Telecommunications}
  \{cgq19,zhoujie18\}@mails.tsinghua.edu.cn, albertyang33@gmail.com, liuzy@tsinghua.edu.cn
}

\renewcommand{\shortauthors}{Cui and Zhou, et al.}
\renewcommand{\algorithmicrequire}{\textbf{Input:}}
\renewcommand{\algorithmicensure}{\textbf{Output:}}
\begin{abstract}
Attributed graph embedding, which learns vector representations from graph topology and node features, is a challenging task for graph analysis.
Recently, methods based on graph convolutional networks (GCNs) have made great progress on this task.
However, existing GCN-based methods have three major drawbacks.
Firstly, our experiments indicate that the entanglement of graph convolutional filters and weight matrices will harm both the performance and robustness.
Secondly, we show that graph convolutional filters in these methods reveal to be special cases of generalized Laplacian smoothing filters, but they do not preserve optimal low-pass characteristics.
Finally, the training objectives of existing algorithms are usually recovering the adjacency matrix or feature matrix, which are not always consistent with real-world applications.
To address these issues, we propose Adaptive Graph Encoder (\ModelName), a novel attributed graph embedding framework. \ModelName\ consists of two modules: (1) To better alleviate the high-frequency noises in the node features, \ModelName\ first applies a carefully-designed Laplacian smoothing filter. 
(2) \ModelName\ employs an adaptive encoder that iteratively strengthens the filtered features for better node embeddings.
We conduct experiments using four public benchmark datasets to validate \ModelName\ on node clustering and link prediction tasks. Experimental results show that \ModelName\ consistently outperforms state-of-the-art graph embedding methods considerably on these tasks.
\blfootnote{$*$ Cheng Yang and Zhiyuan Liu are corresponding authors.}
\end{abstract}


\maketitle

\section{Introduction}

Attributed graphs are graphs with node attributes/features and are widely applied to represent network-structured data in social networks~\cite{hastings2006community}, citation networks~\cite{kipf2016semi}, recommendation systems~\cite{ying2018graph}, etc. For tasks analyzing attributed graphs, including node classification, link prediction and node clustering, plenty of machine learning techniques are developed. However, because of the complex high-dimensional non-Euclidean graph structure and various node features, this task imposes the challenge of jointly capturing structure and feature information on machine learning approaches. 

Representation learning methods on graphs, also known as graph embedding methods, have emerged as general approaches in graph learning area. This kind of approaches aims to learn low-dimensional representations to encode graph structural information. Early graph embedding approaches are based on Laplacian eigenmaps~\cite{newman2006finding}, matrix factorization~\cite{cao2015grarep,yang2015network,li2018community,wang2016semantic}, and random walks~\cite{perozzi2014deepwalk,grover2016node2vec}. However, these methods are also limited because of their shallow architecture. 

\begin{figure}[t]
    \centering
    \includegraphics[width=\linewidth]{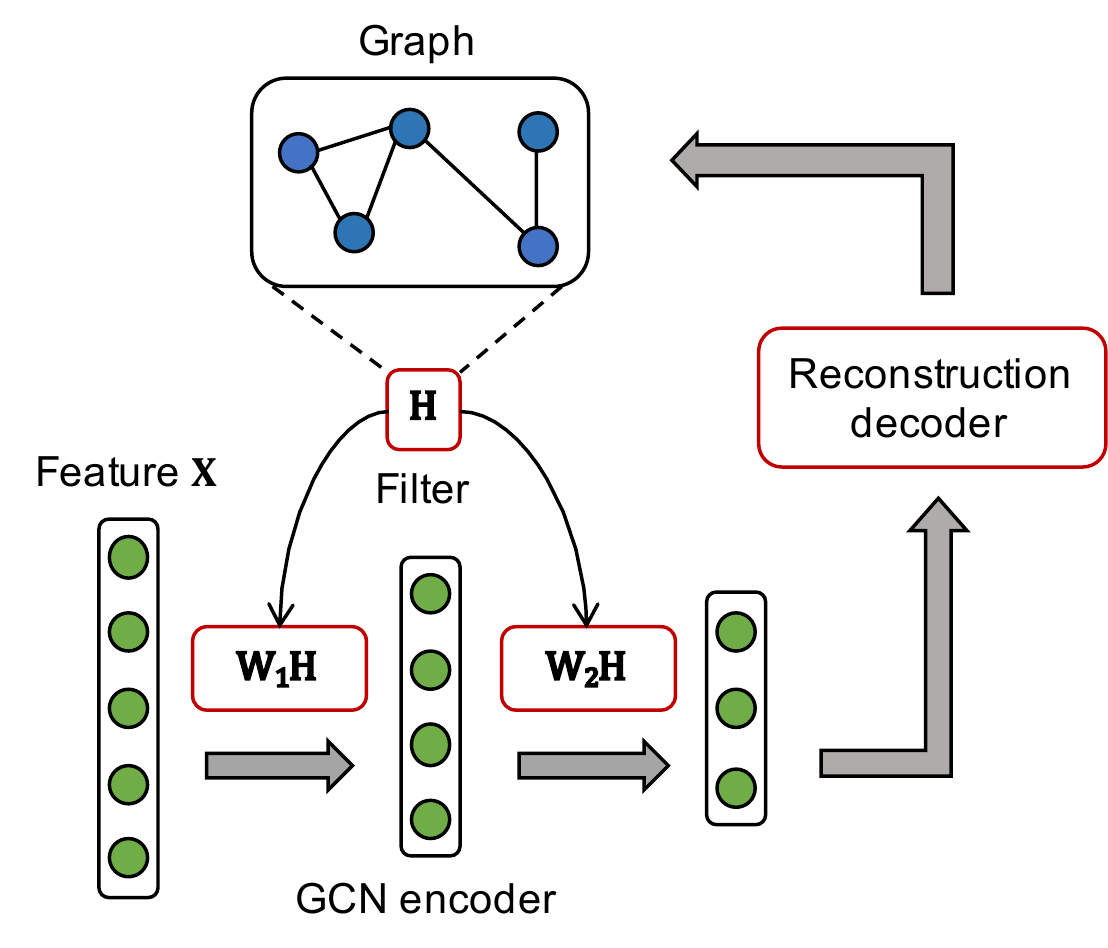}
    \caption{The architecture of graph autoencoder~\cite{kipf2016variational}. The components we argue about are marked in red blocks: Entanglement of the filters and weight matrices, design of the filters, and the reconstruction loss.}
    \label{fig:gae}
\end{figure}

More recently, there has been a surge of approaches that focus on deep learning on graphs. Specifically, approaches from the family of graph convolutional networks (GCNs)~\cite{kipf2016semi} have made great progress in many graph learning tasks~\cite{zhou2018graph} and strengthen the representation power of graph embedding algorithms. 
In this paper, we will study the attributed graph embedding problem, which is one of the most important problems in deep graph learning and GCN-based methods have also made great progress on it.
Among these methods, most of them are based on graph autoencoder (GAE) and variational graph autoencoder (VGAE)~\cite{kipf2016variational}. As shown in Figure~\ref{fig:gae}, they comprise a GCN encoder and a reconstruction decoder. Nevertheless, these GCN-based methods have three major drawbacks: 

Firstly, a GCN encoder consists of multiple graph convolutional layers, and each layer contains a graph convolutional filter ($\mathbf{H}$ in Figure~\ref{fig:gae}), a weight matrix ($\mathbf{W}_1,\mathbf{W}_2$ in Figure~\ref{fig:gae}) and an activation function. However, previous work~\cite{wu2019simplifying} demonstrates that the entanglement of the filters and weight matrices provides no performance gain for semi-supervised graph representation learning, and even harms training efficiency since it deepens the paths of back-propagation. In this work, we further extend this conclusion to unsupervised scenarios by controlled experiments, showing that our disentangled architecture performs better and more robust than entangled models (Section \ref{sec:vs}). 

Secondly, considering the graph convolutional filters, previous research~\cite{li2018deeper} shows in theory that they are actually Laplacian smoothing filters~\cite{taubin1995signal} applied on the feature matrix for low-pass denoising. But we show that existing graph convolutional filters are not optimal low-pass filters since they can not filter out noises in some high-frequency intervals. Thus, they can not reach the best smoothing effect (Section \ref{sec:k}).

Thirdly, we also argue that training objectives of these algorithms (either reconstructing the adjacency matrix~\cite{pan2018adversarially,wang2019attributed} or feature matrix~\cite{wang2017mgae,park2019symmetric}) are not compatible with real-world applications. To be specific, reconstructing adjacency matrix literally sets the adjacency matrix as the ground truth pairwise similarity, while it is not proper for the lack of feature information. 
Recovering the feature matrix, however, will force the model to remember high-frequency noises in features, and thus be inappropriate as well.

Motivated by such observations, we propose Adaptive Graph Encoder (\ModelName), a unified framework for attributed graph embedding.
To disentangle the filters and weight matrices, \ModelName\ consists of two modules: (1) A well-designed non-parametric Laplacian smoothing filter to perform low-pass filtering in order to get smoothed features. (2) An adaptive encoder to learn more representative node embeddings. To replace the reconstruction training objectives, we employ adaptive learning~\cite{chang2017deep} in this step, which selects training samples from the pairwise similarity matrix and finetunes the embeddings iteratively. The code and data are available on \url{https://github.com/thunlp/AGE}.

Our contributions can be summarized as follows:

\begin{itemize}
\item \textit{Analysis}: We make a detailed analysis of the mechanism of graph convolutional filters from the perspective of signal smoothing on graphs and Laplacian smoothing. The analysis helps us design a proper Laplacian smoothing filter to better alleviate high-frequency noises.
\item \textit{Model}: We propose \ModelName, a general model for attributed graph embedding. Our two-fold model disentangles the filters and weight matrices. The filters we adopt preserve the optimal low-pass properties. 
Furthermore, instead of the reconstruction loss, we apply a novel adaptive learning strategy to train node embeddings. 
\item \textit{Experiment}: We conduct extensive experiments on node clustering and link prediction tasks with real-world benchmark datasets. The results demonstrate that \ModelName\ outperforms state-of-the-art attributed graph embedding methods. 
\end{itemize}

\section{Related Work}

\subsection{Conventional Graph Embedding}
Early researches on graph embedding merely focus on finding node similarity with graph structure. Methods based on dimension reduction aim to project the high-dimensional adjacency matrix to low-dimensional latent embedding space. Laplacian eigenmaps~\cite{newman2006finding} and matrix factorization~\cite{cao2015grarep} are two widely used algorithms for these methods. Another line of researches manages to learn node embeddings with a particular objective function.  ~\cite{perozzi2014deepwalk,grover2016node2vec} learn node embeddings by generating random walks and input the sequences into SkipGram model~\cite{le2014distributed}, assuming that similar nodes tend to co-occur in same sequences. 
Other models~\cite{cao2016deep,wang2016structural,tang2015line} can be concluded by an encoder-decoder framework~\cite{hamilton2017representation}, while they differ from model structure and training objectives.

Taking node features into account, there are several works make adjustments to encode structural and content information simultaneously. \cite{yang2015network,li2018community,wang2016semantic} are matrix factorization extensions that add feature-related regularization terms. \cite{chang2009relational,bojchevski2018bayesian} model features as latent variables in Bayesian networks.

\subsection{GCN-based Graph Embedding}
As mentioned in the introduction, due to the strong representation power of graph convolutional networks (GCNs)~\cite{kipf2016semi}, there are several GCN-based approaches for attributed graph embedding and they have achieved state-of-the-art. 
For unsupervised graph embedding that lacks label information, GCN-based methods can be categorized into two groups by their optimization objectives. 

\textbf{Reconstruct the adjacency matrix.} This kind of approaches forces the learned embeddings to recover their localized neighborhood structure.
Graph autoencoder (GAE) and variational graph autoencoder (VGAE)~\cite{kipf2016variational} learn node embeddings by using GCN as the encoder, then decode by inner product with cross-entropy loss. As variants of GAE (VGAE), \cite{pan2018adversarially} exploits adversarially regularized method to learn more robust node embeddings. ~\cite{wang2019attributed} further employs graph attention networks~\cite{velivckovic2017graph} to differentiate the importance of the neighboring nodes to a target node. 

\textbf{Reconstruct the feature matrix.} This kind of models is autoencoders for the node feature matrix while the adjacency matrix merely serves as a filter.
~\cite{wang2017mgae} leverages marginalized denoising autoencoder to disturb the structure information. To build a symmetric graph autoencoder, ~\cite{park2019symmetric} proposes Laplacian sharpening as the counterpart of Laplacian smoothing in the encoder. The authors claim that Laplacian sharpening is a process that makes the reconstructed feature of each node away from the centroid of its neighbors to avoid over-smoothing. However, as we will show in the next section, there exists high-frequency noises in raw node features, which harm the quality of learned embeddings.

\section{Proposed Method}

In this section, we first formalize the embedding task on attributed graphs. Then we present our proposed Adaptive Graph Encoder (\ModelName) algorithm. Specifically, we first design an effective graph filter to perform Laplacian smoothing on node features. Given the smoothed node features, we further develop a simple node representation learning module based on adaptive learning~\cite{chang2017deep}. Finally, the learned node embeddings are used for downstream tasks such as node clustering and link prediction.

\subsection{Problem Formalization}
Given an attributed graph $\mathcal{G}=(\mathcal{V}, \mathcal{E}, \mathbf{X})$, where $\mathcal{V}=\{v_1, v_2, \cdots , v_n\}$ is the vertex set with $n$ nodes in total, $\mathcal{E}$ is the edge set, and $\mathbf{X}=[\mathbf{x}_1, \mathbf{x}_2. \cdots, \mathbf{x}_n]^T$ is the feature matrix. The topology structure of graph $\mathcal{G}$ can be denoted by an adjacency matrix $\mathbf{A}=\{a_{ij}\} \in \mathbb{R}^{n\times n}$, where $a_{ij}=1$ if $(v_i,v_j)\in \mathcal{E}$, indicating there is an edge from node $v_i$ to node $v_j$. $\mathbf{D}=\text{diag}(d_1, d_2,\cdots,d_n)\in\mathbb{R}^{n\times n}$ denotes the degree matrix of $\mathbf{A}$, where $d_i = \sum_{v_j\in\mathcal{V}}a_{ij}$ is the degree of node $v_i$. The graph Laplacian matrix is defined as $\mathbf{L}=\mathbf{D-A}$. 

The purpose of attributed graph embedding is to map nodes to low-dimensional embeddings. We take $\mathbf{Z}$ as the embedding matrix and the embeddings should preserve both the topological structure and feature information of graph $\mathcal{G}$.

For downstream tasks, we consider node clustering and link prediction.
The node clustering task aims to partition the nodes into $m$ disjoint groups $\{G_1, G_2, \cdots, G_m\}$, where similar nodes should be in the same group. The link prediction task requires the model to predict whether there is a potential edge existing between two given nodes.

\subsection{Overall Framework}

\begin{figure}[t]
    \centering
    \includegraphics[width=\linewidth]{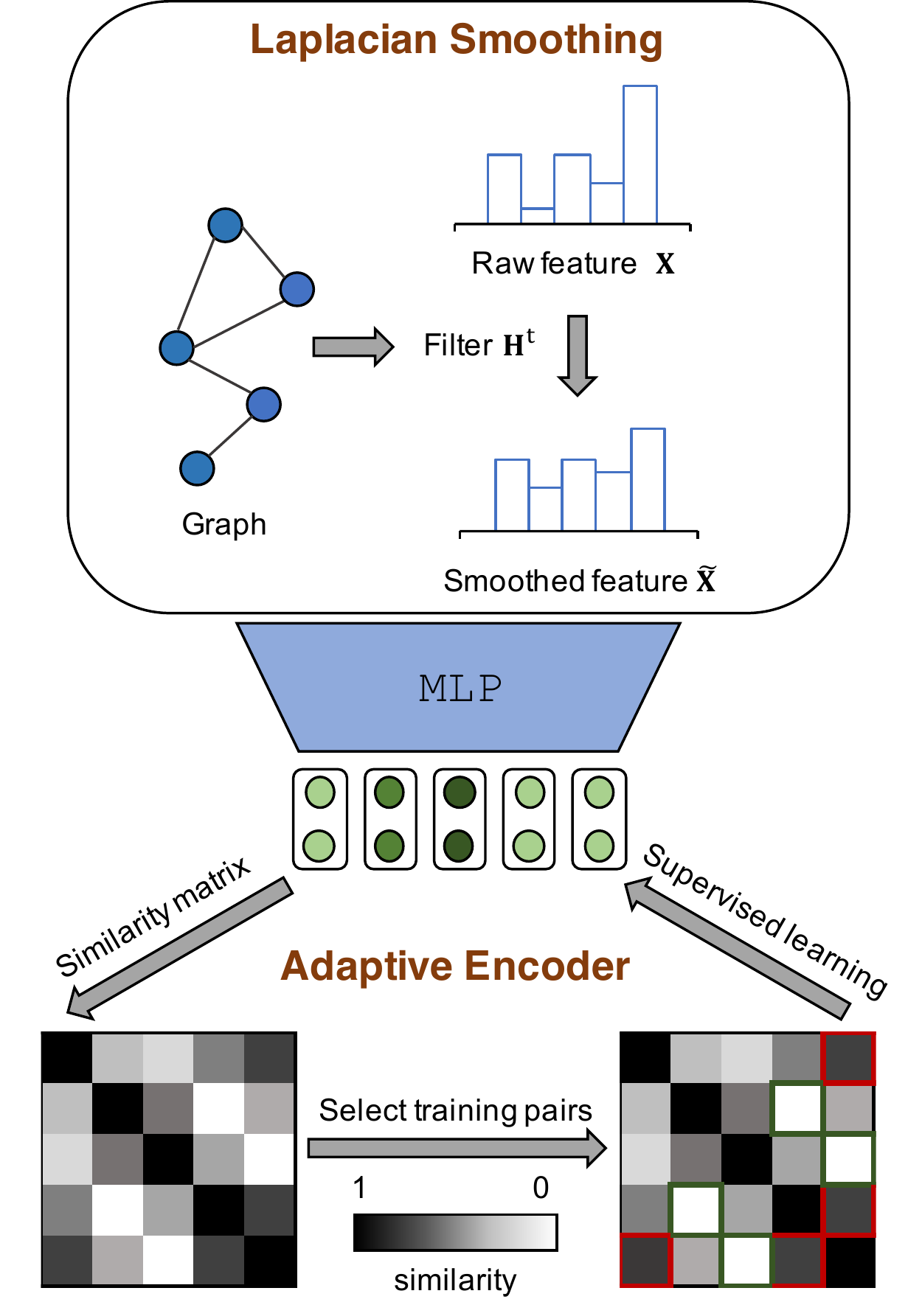}
    \caption{Our \ModelName\ framework. Given the raw feature matrix $\mathbf{X}$, we first perform $t$-layer Laplacian smoothing using filter $\mathbf{H}^t$ to get the smoothed feature matrix $\tilde{\mathbf{X}}$ (\textbf{Top}). Then the node embeddings are encoded by the adaptive encoder which utilizes the adaptive learning strategy: (1) Calculate the pairwise node similarity matrix. (2) Select positive and negative training samples of high confidence (red and green squares). (3) Train the encoder by a supervised loss (\textbf{Bottom}). }
    \label{model}
\end{figure}

The framework of our model is shown in Figure~\ref{model}. It consists of two parts: a Laplacian smoothing filter and an adaptive encoder.
\begin{itemize}
    \item \textbf{Laplacian Smoothing Filter}: The designed filter $\mathbf{H}$ serves as a low-pass filter to denoise the high-frequency components of the feature matrix $\mathbf{X}$. The smoothed feature matrix $\tilde{\mathbf{X}}$ is taken as input of the adaptive encoder.
    \item \textbf{Adaptive Encoder}: To get more representative node embeddings, this module builds a training set by adaptively selecting node pairs which are highly similar or dissimilar. Then the encoder is trained in a supervised manner.
\end{itemize}
After the training process, the learned node embedding matrix $\mathbf{Z}$ is used for downstream tasks. 

\subsection{Laplacian Smoothing Filter}

The basic assumption for graph learning is that nearby nodes on the graph should be similar, thus node features are supposed to be \textit{smooth} on the graph manifold. In this section, we first explain what \textit{smooth} means. Then we give the definition of the generalized Laplacian smoothing filter and show that it is a smoothing operator. Finally, we answer how to design an optimal Laplacian smoothing filter.

\subsubsection{Analysis of Smooth Signals} 
We start with interpreting \textit{smooth} from the perspective of graph signal processing. Take $\mathbf{x}\in \mathbb{R}^n$ as a graph signal where each node is assigned with a scalar. Denote the filter matrix as $\mathbf{H}$. 
To measure the smoothness of graph signal $\mathbf{x}$, we can calculate the \textit{Rayleigh quotient}~\cite{horn2012matrix} over the graph Laplacian $\mathbf{L}$ and $\mathbf{x}$:
\begin{equation}
    \label{eq:ray}
    R(\mathbf{L}, \mathbf{x})=\frac{\mathbf{x^\intercal L x}}{\mathbf{x^\intercal x}}=
    \frac{\sum_{(i,j)\in\mathcal{E}} (x_i-x_j)^2}{\sum_{i\in\mathcal{V}}x_i^2}.
\end{equation}
This quotient is actually the normalized variance score of $\mathbf{x}$. As stated above, \textit{smooth} signals should assign similar values on neighboring nodes. Consequently, signals with lower \textit{Rayleigh quotient} are assumed to be smoother.

Consider the eigendecomposition of graph Laplacian $\mathbf{L}=\mathbf{U\Lambda U}^{-1}$, where $\mathbf{U}\in\mathbb{R}^{n\times n}$ comprises eigenvectors and $\mathbf{\Lambda}=\text{diag}(\lambda_1,\lambda_2,\cdots,\lambda_n)$ is a diagonal matrix of eigenvalues. Then the smoothness of eigenvector $\mathbf{u}_i$ is given by
\begin{equation}
    R(\mathbf{L}, \mathbf{u}_i)=\frac{\mathbf{u}_i^\intercal \mathbf{L u}_i}{\mathbf{u}_i^\intercal \mathbf{u}_i}=\lambda_i.
    \label{eq:smooth}
\end{equation}
Eq.~(\ref{eq:smooth}) indicates that smoother eigenvectors are associated with smaller eigenvalues, which means lower frequencies. Thus we decompose signal $\mathbf{x}$ on the basis of $\mathbf{L}$ based on Eq.~(\ref{eq:ray}) and Eq.~(\ref{eq:smooth}):
\begin{equation}
    \mathbf{x} = \mathbf{Up} = \sum_{i=1}^n p_i\mathbf{u}_i.
\end{equation}
where $p_i$ is the coefficient of eigenvector $\mathbf{u}_i$. Then the smoothness of $\mathbf{x}$ is actually
\begin{equation}
    R(\mathbf{L}, \mathbf{x})=\frac{\mathbf{x^\intercal L x}}{\mathbf{x^\intercal x}}=\frac{\sum_{i=1}^n p_i^2\lambda_i}{\sum_{i=1}^n p_i^2}.
\end{equation}

Therefore, to get smoother signals, the goal of our filter is filtering out high-frequency components while preserving low-frequency components. Because of its high computational efficiency and convincing performance, Laplacian smoothing filters~\cite{taubin1995signal} are often utilized for this purpose.

\subsubsection{Generalized Laplacian Smoothing Filter} 
As stated by \cite{taubin1995signal}, the generalized Laplacian smoothing filter is defined as 
\begin{equation}
    \label{eq:old-filter}
    \mathbf{H}=\mathbf{I}-k\mathbf{L},
\end{equation}
where $k$ is real-valued. Employ $\mathbf{H}$ as the filter matrix, the filtered signal $\tilde{\mathbf{x}}$ is present by
\begin{equation}
    \tilde{\mathbf{x}} = \mathbf{Hx} = \mathbf{U}(\mathbf{I}-k\mathbf{\Lambda})\mathbf{U}^{-1} \mathbf{Up}=\sum_{i=1}^n(1-k\lambda_i)p_i\mathbf{u}_i=\sum_{i=1}^np\prime_i\mathbf{u}_i.
\end{equation}
Hence, to achieve low-pass filtering, the frequency response function $1-k\lambda$ should be a decrement and non-negative function.
Stacking up $t$ Laplacian smoothing filters, we denote the filtered feature matrix $\tilde{\textbf{X}}$ as
\begin{equation}
    \label{eq:sx}
    \tilde{\mathbf{X}} = \mathbf{H}^t\mathbf{X}.
\end{equation}
Note that the filter is non-parametric at all.

\subsubsection{The Choice of $k$} 
\label{sec:k}
In practice, with the renormalization trick $\tilde{\mathbf{A}} = \mathbf{I}+\mathbf{A}$, we employ the symmetric normalized graph Laplacian
\begin{equation}
    \label{eq:lsym}
    \tilde{\mathbf{L}}_{sym}=\tilde{\mathbf{D}}^{-\frac{1}{2}}\tilde{\mathbf{L}}\tilde{\mathbf{D}}^{-\frac{1}{2}},
\end{equation}
 where $\tilde{\mathbf{D}}$ and $\tilde{\mathbf{L}}$ are degree matrix and Laplacian matrix corresponding to $\tilde{\mathbf{A}}$. Then the filter becomes
 \begin{equation}
    \label{eq:filter}
    \mathbf{H}=\mathbf{I}-k\tilde{\mathbf{L}}_{sym}.
\end{equation}
 Notice that if we set $k=1$, the filter becomes the GCN filter. 

For selecting optimal $k$, the distribution of eigenvalues $\tilde{\mathbf{\Lambda}}$ (obtained from the decomposition of  $\tilde{\mathbf{L}}_{sym}=\mathbf{\tilde{U} \tilde{\Lambda} \tilde{U}^{-1}}$) should be carefully discovered. 

The smoothness of $\tilde{\mathbf{x}}$ is 
\begin{equation}
    R(\mathbf{L}, \tilde{\mathbf{x}})=\frac{\tilde{\mathbf{x}}^\intercal \mathbf{L} \tilde{\mathbf{x}}}{\tilde{\mathbf{x}}^\intercal \tilde{\mathbf{x}}}=\frac{\sum_{i=1}^n p\prime_i^2\lambda_i}{\sum_{i=1}^n p\prime_i^2}.
\end{equation}
Thus $p\prime_i^2$ should decrease as $\lambda_i$ increases. We denote the maximum eigenvalue as $\lambda_{max}$. Theoretically, if $k>1/\lambda_{max}$, the filter is not low-pass in the $(1/k, \lambda_{max}]$ interval because $p\prime_i^2$ increases in this interval; Otherwise, if $k<1/\lambda_{max}$, the filter can not denoise all the high-frequency components. Consequently, $k=1/\lambda_{max}$ is the optimal choice.

It has been proved that the range of Laplacian eigenvalues is between 0 and 2~\cite{chung1997spectral}, hence GCN filter is not low-pass in the $(1,2]$ interval. Some work~\cite{wang2019attributed} accordingly chooses $k=1/2$. However, our experiments show that after renormalization, the maximum eigenvalue $\lambda_{max}$ will shrink to around $3/2$, which makes $1/2$ not optimal as well.
In experiments, we calculate $\lambda_{max}$ for each dataset and set $k=1/\lambda_{max}$. We further analyse the effects of different $k$ values (Section \ref{sec:kexp}).

\subsection{Adaptive Encoder}

Filtered by $t$-layer Laplacian smoothing, the output features are smoother and preserve abundant attribute information. 


To learn better node embeddings from the smoothed features, we need to find an appropriate unsupervised optimization objective. To this end, we manage to utilize pairwise node similarity inspired by Deep Adaptive Learning~\cite{chang2017deep}.
For attributed graph embedding task, the relationship between two nodes is crucial, which requires the training targets to be suitable similarity measurements. GAE-based methods usually choose the adjacency matrix as true labels of node pairs. However, we argue that the adjacency matrix only records one-hop structure information, which is insufficient. Meanwhile, we address that the similarity of smoothed features or trained embeddings are more accurate since they incorporate structure and features together. To this end, we adaptively select node pairs of high similarity as positive training samples, while those of low similarity as negative samples.

Given filtered node features $\tilde{\mathbf{X}}$, the node embeddings are encoded by linear encoder $f$:
\begin{equation}
    \label{eq:z}
    \mathbf{Z} = f(\tilde{\mathbf{X}}; \mathbf{W})=\tilde{\mathbf{X}}\mathbf{W},
\end{equation}
where $\mathbf{W}$ is the weight matrix. We then scale the embeddings to the $[0,1]$ interval by min-max scaler for variance reduction. To measure the pairwise similarity of nodes, we utilize cosine function to implement our similarity metric. The similarity matrix $\mathbf{S}$ is given by
\begin{equation}
    \label{eq:s}
    \mathbf{S}=\frac{\mathbf{ZZ}^\intercal}{ \Vert\mathbf{Z}\Vert_2^2}.
\end{equation}

Next, we describe our training sample selection strategy in detail.

\subsubsection{Training Sample Selection} After calculating the similarity matrix, we rank the pairwise similarity sequence in the descending order. Here $r_{ij}$ is the rank of node pair $(v_i, v_j)$. Then we set the maximum rank of positive samples as $r_{pos}$ and the minimum rank of negative samples as $r_{neg}$. Therefore, the generated label of node pair $(v_i, v_j)$ is
\begin{equation}
    \label{eq:label}
    l_{ij} = 
    \begin{cases}
    1 & r_{ij} \le r_{pos}\\
    0 & r_{ij} > r_{neg}\\
    \text{None} & \text{otherwise}\\
    \end{cases}.
\end{equation}
In this way, a training set with $r_{pos}$ positive samples and $n^2-r_{neg}$ negative samples is constructed. Specially, for the first time we construct the training set, since the encoder is not trained, we directly employ the smoothed features for initializing $\mathbf{S}$:
\begin{equation}
    \label{eq:s0}
    \mathbf{S}=\frac{\mathbf{\tilde{X}\tilde{X}}^\intercal}{\Vert\mathbf{\tilde{X}}\Vert_2^2}.
\end{equation}

After construction of the training set, we can train the encoder in a supervised manner. In real-world graphs, there are always far more dissimilar node pairs than positive pairs, so we select more than $r_{pos}$ negative samples in the training set. To balance positive/negative samples, we randomly choose $r_{pos}$ negative samples in every epoch. The balanced training set is denoted by $\mathcal{O}$. Accordingly, our cross entropy loss is given by
\begin{equation}
\label{eq:loss}
\mathcal{L}=\sum_{(v_i, v_j)\in\mathcal{O}} -l_{ij}\log(s_{ij})-(1-l_{ij})\log(1-s_{ij}).
\end{equation}

\subsubsection{Thresholds Update} Inspired by the idea of curriculum learning~\cite{bengio2009curriculum}, we design a specific update strategy for $r_{pos}$ and $r_{neg}$ to control the size of training set. At the beginning of training process, more samples are selected for the encoder to find rough cluster patterns. After that, samples with higher confidence are remained for training, forcing the encoder to capture refined patterns. In practice, $r_{pos}$ decreases while $r_{neg}$ increases linearly as the training procedure goes on. We set the initial threshold as $r_{pos}^{st}$ and $r_{neg}^{st}$, together with the final threshold as $r_{pos}^{ed}$ and $r_{neg}^{ed}$. We have $r_{pos}^{ed}\le r_{pos}^{st}$ and $r_{neg}^{ed} \ge r_{neg}^{st}$. Suppose the thresholds are updated $T$ times, we present the update strategy as
\begin{equation}
    \label{eq:pos}
    r^{\prime}_{pos} = r_{pos} + \frac{r_{pos}^{ed}-r_{pos}^{st}}{T},
\end{equation}
\begin{equation}
    \label{eq:neg}
    r^{\prime}_{neg} = r_{neg} + \frac{r_{neg}^{ed}-r_{neg}^{st}}{T}.
\end{equation}

As the training process goes on, every time the thresholds are updated, we reconstruct the training set and save the embeddings. For node clustering, we perform Spectral Clustering~\cite{ng2002spectral} on the similarity matrices of saved embeddings, and select the best epoch by Davies–Bouldin index~\cite{davies1979cluster} (DBI), which measures the clustering quality without label information. For link prediction, we select the best performed epoch on validation set.
Algorithm~\ref{alg} presents the overall procedure of computing the embedding matrix $\mathbf{Z}$.

\begin{algorithm}

    \caption{Adaptive Graph Encoder}
    \begin{algorithmic}[1]
    \label{alg}
        \REQUIRE  Adjacency matrix $\mathbf{A}$, feature matrix $\mathbf{X}$, filter layer number $t$, iteration number $max\_iter$ and threshold update times $T$
        \ENSURE Node embedding matrix $\mathbf{Z}$
        \STATE Obtain graph Laplacian $\tilde{\mathbf{L}}_{sym}$ from Eq.~(\ref{eq:lsym});
        \STATE $k \gets 1/\lambda_{max}$;
        \STATE Get filter matrix $\mathbf{H}$ from Eq.~(\ref{eq:filter});
        \STATE Get smoothed feature matrix $\tilde{\mathbf{X}}$ from Eq.~(\ref{eq:sx});
        \STATE Initialize similarity matrix $\mathbf{S}$ and training set $\mathcal{O}$ by Eq.~(\ref{eq:s0});
        \FOR {$iter=1$ \TO $max\_iter$}
        \STATE Compute $\mathbf{Z}$ with Eq.~(\ref{eq:z});
        \STATE Train the adaptive encoder with loss in Eq.~(\ref{eq:loss});
        \IF{$iter\mod (max\_iter/T) == 0$}
        \STATE Update thresholds with Eq.~(\ref{eq:pos}) and ~(\ref{eq:neg});
        \STATE Calculate the similarity matrix $\mathbf{S}$ with Eq.~(\ref{eq:s});
        \STATE Select training samples from $\mathbf{S}$ by Eq.~(\ref{eq:label});
        \ENDIF
        \ENDFOR
        
    \end{algorithmic}
\end{algorithm}

\section{Experimental Settings}

We evaluate the benefits of \ModelName\ against a number of state-of-the-art graph embedding approaches on node clustering and link prediction tasks. In this section, we introduce our benchmark datasets, baseline methods, evaluation metrics, and parameter settings.

\subsection{Datasets}
We conduct node clustering and link prediction experiments on four widely used network datasets (Cora, Citeseer, Pubmed~\cite{sen2008collective} and Wiki~\cite{yang2015network}). Features in Cora and Citeseer are binary word vectors, while in Wiki and Pubmed, nodes are associated with tf-idf weighted word vectors. The statistics of the four datasets are shown in Table~\ref{tab:data}.
\begin{table}[ht]
    \centering
    \caption{Dataset statistics}
    \begin{tabular}{l|cccc}
    \toprule
        Dataset & \# Nodes & \# Edges & \# Features & \# Classes \\
    \midrule
       Cora & 2,708 & 5,429 & 1,433 & 7 \\
       Citeseer & 3,327 & 4,732 & 3,703 & 6\\
       Wiki & 2,405 & 17,981 & 4,973 & 17\\
       Pubmed & 19,717 & 44,338 & 500 & 3\\
     \bottomrule
    \end{tabular}
    \label{tab:data}
\end{table}

\begin{table*}[!htbp]
	\centering
	\caption{Experimental results of node clustering.}\label{tab:mainexp}
	\begin{tabular}{l|c|ccc|ccc|ccc|ccc}
		\toprule
		 \multirow{2}{*}{Methods} & \multirow{2}{*}{Input} & \multicolumn{3}{|c|}{Cora} & \multicolumn{3}{|c|}{Citeseer} & \multicolumn{3}{|c|}{Wiki} & \multicolumn{3}{|c}{Pubmed} \\
		 \cmidrule{3-14}
		  & & ACC & NMI & ARI & ACC & NMI & ARI & ACC & NMI & ARI & ACC & NMI & ARI \\
		\midrule
		Kmeans & F & 0.503 & 0.317 & 0.244 & 0.544 & 0.312 & 0.285 & 0.417 & 0.440 & 0.151 & 0.580 & 0.278 & 0.246 \\
		Spectral-F & F & 0.347 & 0.147 & 0.071 & 0.441 & 0.203 & 0.183 & 0.491 & 0.464 & 0.254 & 0.602 & 0.309 & 0.277 \\
		\midrule
		Spectral-G & G & 0.342 & 0.195 & 0.045 & 0.259 & 0.118 & 0.013 & 0.236 & 0.193 & 0.017 & 0.528 & 0.097 & 0.062 \\
		DeepWalk & G & 0.484 & 0.327 & 0.243 & 0.337 & 0.089 & 0.092 & 0.385 & 0.324 & 0.173 & 0.543 & 0.102 & 0.088 \\
		\midrule
		TADW & F\&G & 0.560 & 0.441 & 0.332 & 0.455 & 0.291 & 0.228 & 0.310 & 0.271 & 0.045 & 0.511 & 0.244 & 0.217 \\
		GAE & F\&G & 0.611 & 0.482 & 0.302 & 0.456 & 0.221 & 0.191 & 0.379 & 0.345 & 0.189 & 0.632 & 0.249 & 0.246 \\
		VGAE & F\&G & 0.592 & 0.408 & 0.347 & 0.467 & 0.261 & 0.206 & 0.451 & 0.468 & 0.263 & 0.619 & 0.216 & 0.201 \\
		MGAE & F\&G & 0.681 & 0.489 & 0.436 & 0.669 & 0.416 & 0.425 & 0.529 & \underline{0.510} & 0.379 & 0.593 & 0.282 & 0.248 \\
		ARGA & F\&G & 0.640 & 0.449 & 0.352 & 0.573 & 0.350 & 0.341 & 0.381 & 0.345 & 0.112 & 0.681 & 0.276 & 0.291 \\
		ARVGA & F\&G & 0.638 & 0.450 & 0.374 & 0.544 & 0.261 & 0.245 & 0.387 & 0.339 & 0.107 & 0.513 & 0.117 & 0.078 \\
		AGC & F\&G & 0.689 & 0.537 & 0.486 & 0.670 & 0.411 & 0.419 & 0.477 & 0.453 & 0.343 & \underline{0.698} & 0.316 & 0.319 \\
		DAEGC & F\&G & 0.704 & 0.528 & 0.496 & 0.672 & 0.397 & 0.410 & 0.482 & 0.448 & 0.331 & 0.671 & 0.266 & 0.278\\
		GALA & F\&G & \underline{0.746} & \underline{0.577} & \underline{0.532} & \underline{0.693} & \underline{0.441} & \underline{0.446} & \underline{0.545} & 0.504 & \underline{0.389} & 0.694 & \underline{\bf0.327} & \underline{0.321}\\
		\midrule
		LS & F\&G & 0.638 & 0.493 & 0.373 & 0.677 & 0.419 & 0.433 & 0.515 & 0.534 & 0.317 & 0.656 & 0.300 & 0.315\\
		LS+RA & F\&G & 0.742 & 0.580 & 0.545 & 0.658 & 0.410 & 0.403 & 0.552 & 0.566 & 0.382 & 0.652 & 0.291 & 0.301\\
		LS+RX & F\&G & 0.647 & 0.479 &0.423 &0.674 &0.416 &0.424 &0.553 &0.543 &0.365 &0.645 &0.285 & 0.251\\ 

		\ModelName & F\&G & \bf 0.768 & \bf0.607 & \bf0.565 & \bf0.702 & \bf0.448 & \bf0.457 & \bf0.612 & \bf0.597 & \bf0.440 & \bf0.711 & 0.316 & \bf0.334\\
        \bottomrule
	\end{tabular}
\end{table*}

\subsection{Baseline Methods}
For attributed graph embedding methods, we include 5 baseline algorithms in our comparisons:

GAE and VGAE~\cite{kipf2016variational} combine graph convolutional networks with the (variational) autoencoder for representation learning. 

ARGA and ARVGA~\cite{pan2018adversarially} add adversarial constraints to GAE and VGAE respectively, enforcing the latent representations to match a prior distribution for robust node embeddings. 

GALA~\cite{park2019symmetric} proposes a symmetric graph convolutional autoencoder recovering the feature matrix. The encoder is based on Laplacian smoothing while the decoder is based on Laplacian sharpening.

On the node clustering task, we compare our model with 8 more algorithms. The baselines can be categorized into three groups: 

{\textbf{(1) Methods using features only.}}
Kmeans~\cite{lloyd1982least} and Spectral Clustering~\cite{ng2002spectral} are two traditional clustering algorithms. Spectral-F takes the cosine similarity of node features as input.

{\textbf{(2) Methods using graph structure only.}}
Spectral-G is Spectral Clustering with the adjacency matrix as the input similarity matrix. 
DeepWalk~\cite{perozzi2014deepwalk} learns node embeddings by using SkipGram on generated random walk paths on graphs. 

{\textbf{(3) Methods using both features and graph.}}
TADW~\cite{yang2015network} interprets DeepWalk as matrix factorization and incorporates node features under the DeepWalk framework. 
MGAE~\cite{wang2017mgae} is a denoising marginalized graph autoencoder. Its training objective is reconstructing the feature matrix.
AGC~\cite{zhang2019attributed} exploits high-order graph convolution to filter node features. The number of graph convolution layers are selected for different datasets.
DAEGC~\cite{wang2019attributed} employs graph attention network to capture the importance of the neighboring nodes, then co-optimize reconstruction loss and KL-divergence-based clustering loss.

For representation learning algorithms including DeepWalk, TADW, GAE and VGAE which do not specify on the node clustering problem, we apply Spectral Clustering on their learned representations. For other works that conduct experiments on benchmark datasets, the original results in the papers are reported.

\textbf{\ModelName\ variants}.
We consider 4 variants of \ModelName\ to compare various optimization objectives. The Laplacian smoothing filters in these variants are the same, while the encoder of LS+RA aims at reconstructing the adjacency matrix. LS+RX, respectively, reconstructs the feature matrix. LS only preserves the Laplacian smoothing filter, the smoothed features are taken as node embeddings. \ModelName\ is our proposed model with adaptive learning. 

\subsection{Evaluation Metrics \& Parameter Settings}
To measure the performance of node clustering methods, we employ three metrics: Accuracy (ACC), Normalized Mutual Information (NMI), and Adjusted Rand Index (ARI)~\cite{gan2007data}. For link prediction, we partition the datasets following GAE, and report Area Under Curve (AUC) and Average Precision (AP) scores. For all the metrics, a higher value indicates better performance.

For the Laplacian smoothing filter, we find the maximum eigenvalues of the four datasets are all around $3/2$. Thus we set $k=2/3$ universally. For the adaptive encoder, we train the MLP encoder for 400 epochs with a 0.001 learning rate by the Adam optimizer~\cite{kingma2014adam}. The encoder consists of a single 500-dimensional embedding layer, and we update the thresholds every 10 epochs. We tune other hyperparameters including Laplacian smoothing filter layers $t$, $r_{pos}^{st}$, $r_{pos}^{ed}$, $r_{neg}^{st}$ and $r_{neg}^{ed}$ based on DBI. The detailed hyperparameter settings are reported in Appendix. 

\begin{figure*}[ht]
    \centering
    \begin{subfigure}[h]{0.24\linewidth}
    \includegraphics[width=\linewidth]{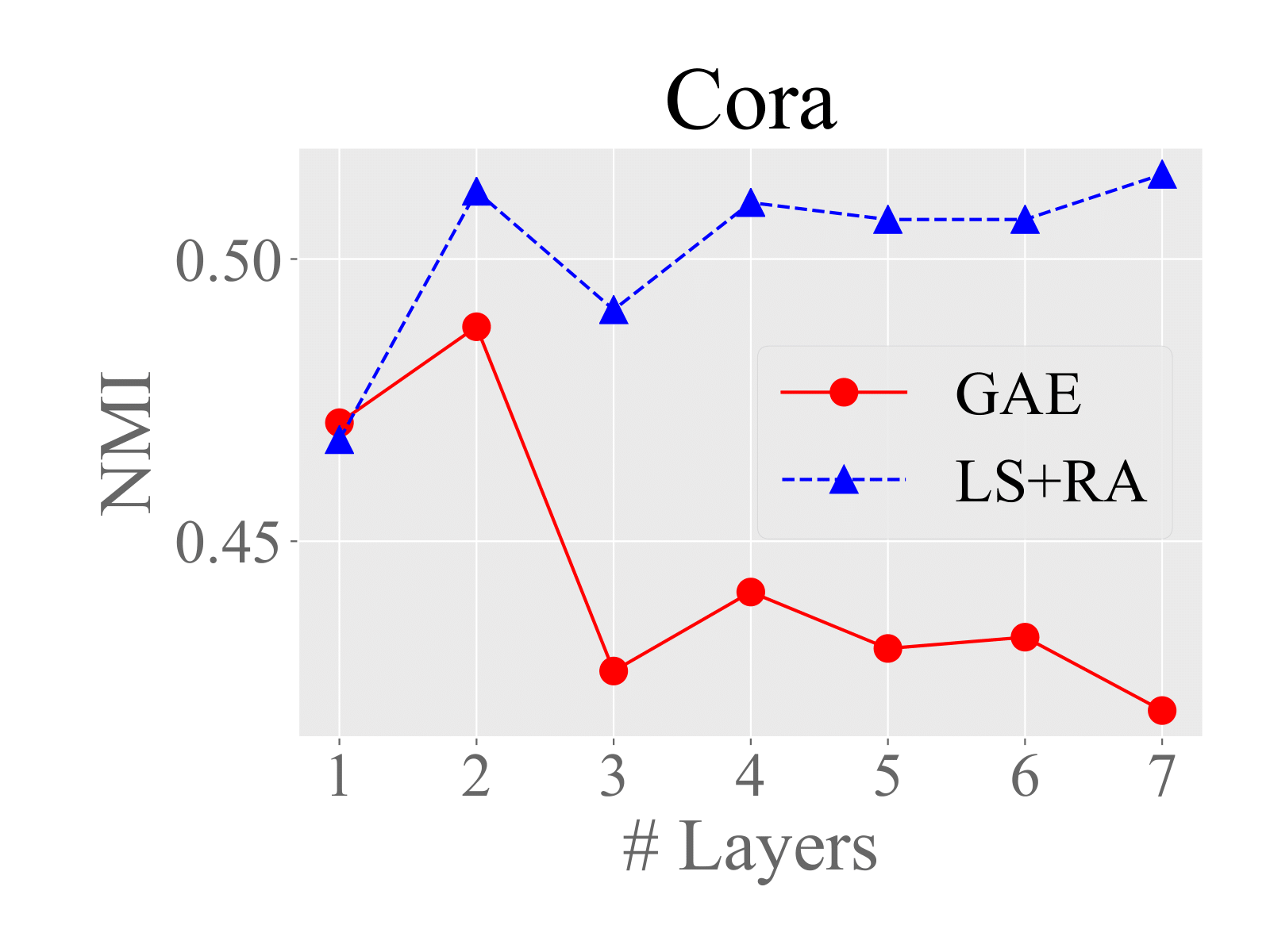}
    \end{subfigure}
    \begin{subfigure}[h]{0.24\linewidth}
    \includegraphics[width=\linewidth]{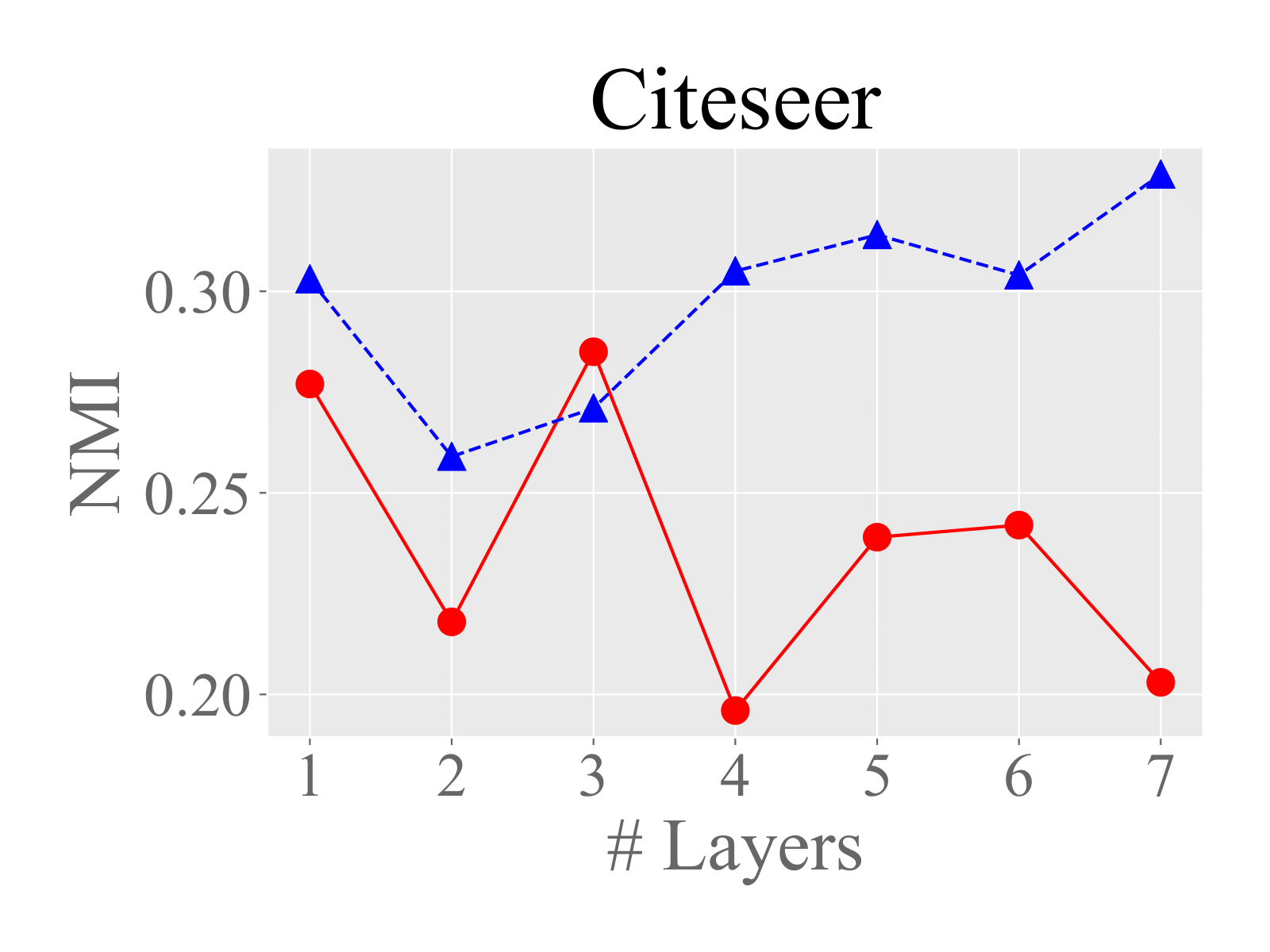}
    \end{subfigure}
    \begin{subfigure}[h]{0.24\linewidth}
    \includegraphics[width=\linewidth]{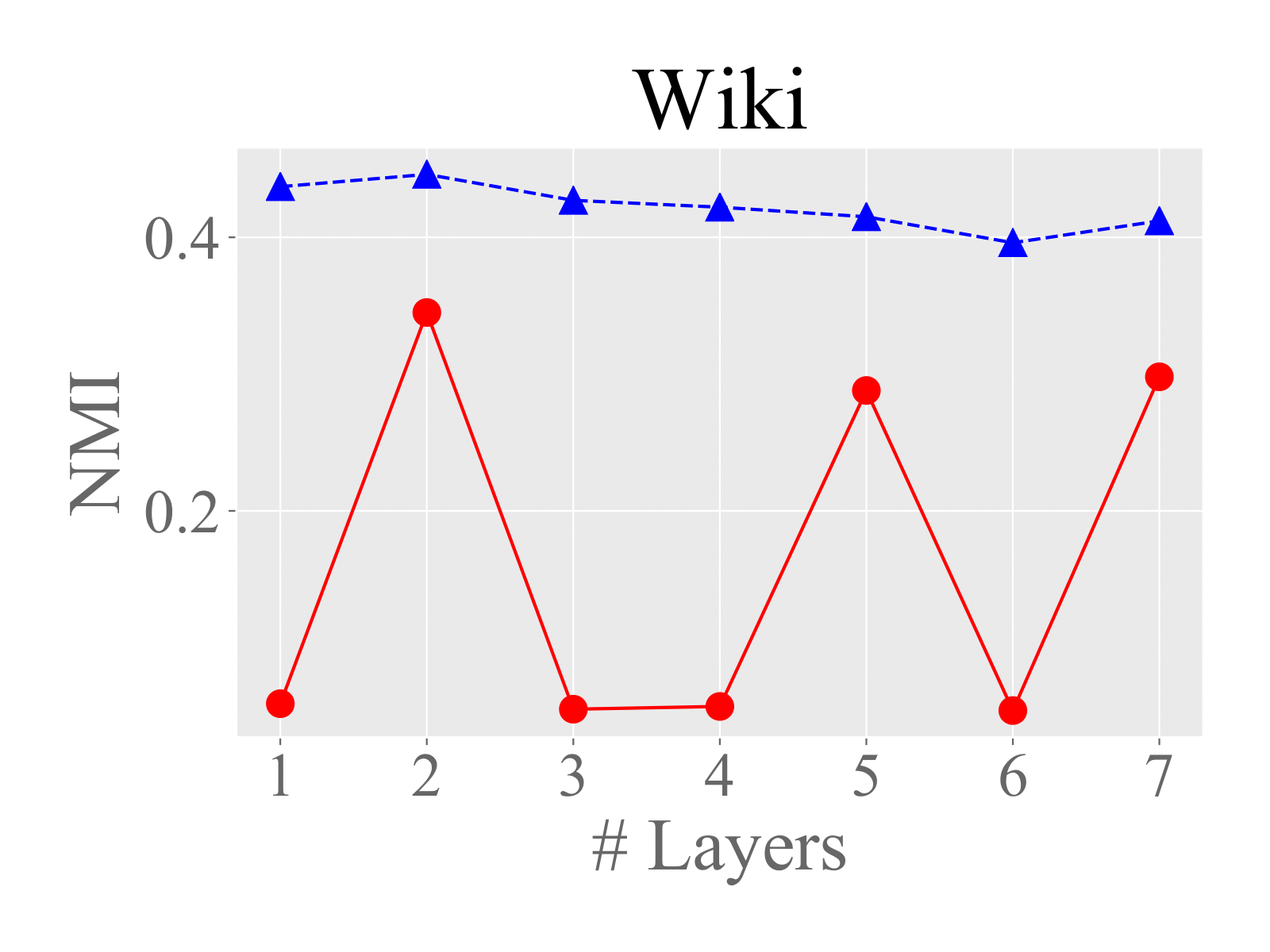}
    \end{subfigure}
    \begin{subfigure}[h]{0.24\linewidth}
    \includegraphics[width=\linewidth]{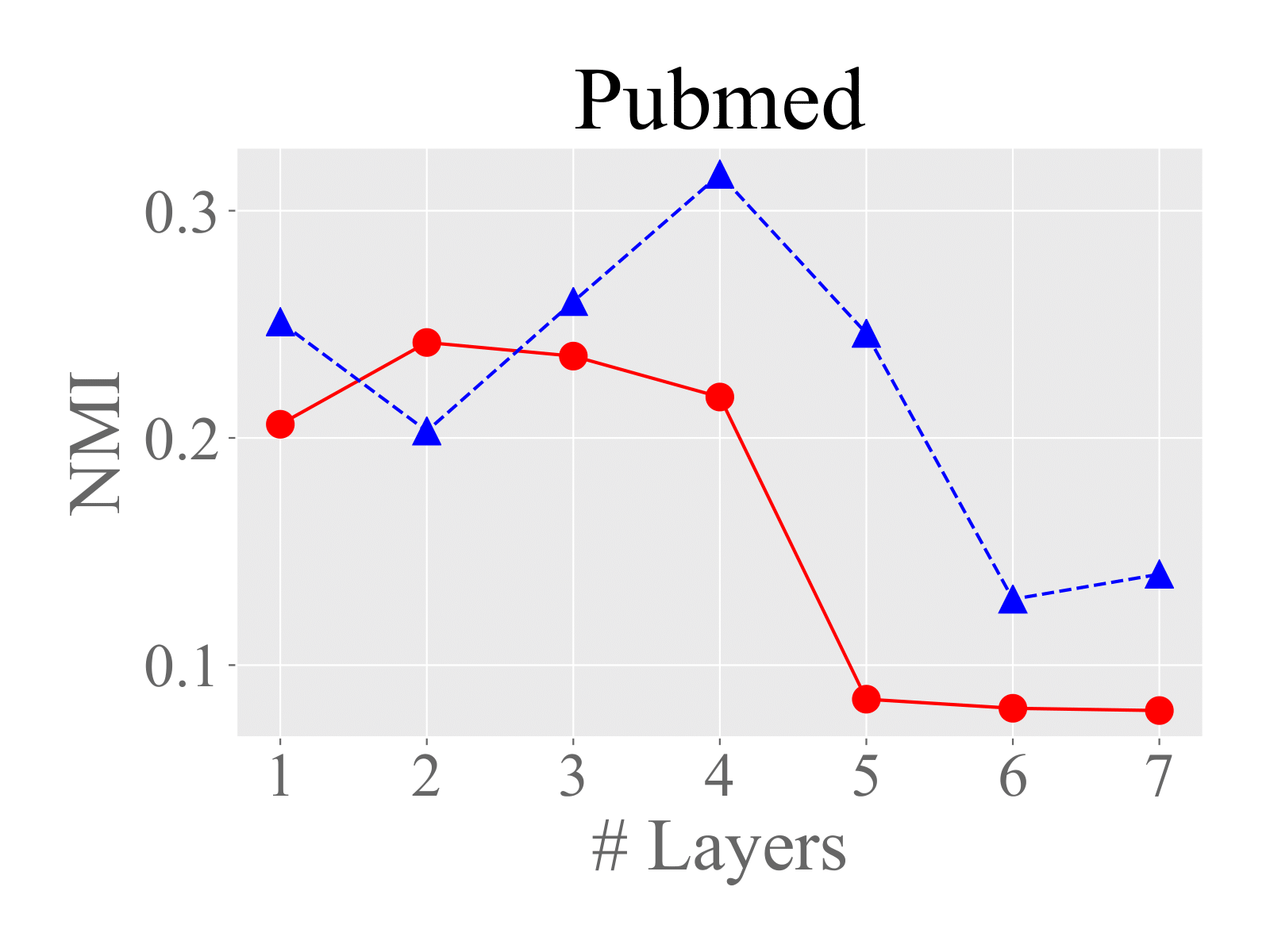}
    \end{subfigure}
    \caption{Controlled experiments comparing GAE and LS+RA}
    \label{fig:vs}
\end{figure*}

\section{Experimental Results}
In this section, we show and analyse the results of our experiments. Besides the main experiments, we also conduct auxiliary experiments to answer the following hypotheses:

\textbf{H1}: Entanglement of the filters and weight matrices has no improvement for embedding quality.

\textbf{H2}: Our adaptive learning strategy is effective compared to reconstruction losses, and each mechanism has its own contribution.

\textbf{H3}: $k=1/\lambda_{max}$ is the optimal choice for Laplacian smoothing filters.

\subsection{Node Clustering Results}

The node clustering results are presented in Table~\ref{tab:mainexp}, where \textbf{bold} and \underline{underlined} values indicate the highest scores in all methods and all baselines respectively. Our observations are as follows:

Algorithms using both feature and graph information usually achieve better performance than methods leveraging information from single source. This investigation demonstrates that features and graph structure contribute to clustering from different perspectives.

\ModelName\ shows superior performance to baseline methods by a considerable margin, especially on Cora and Wiki datasets. Competing with the strongest baseline GALA, our model outperforms it by 2.95\%, 5.20\% and 6.20\% on Cora, by 12.29\%, 18.45\% and 13.11\% on Wiki with respect to ACC, NMI and ARI. Such results show strong evidence advocating our proposed framework. For Citeseer and Pubmed, we give further analysis in section \ref{sec:kexp}.

Compared with GCN-based methods, \ModelName\ has simpler mechanisms than those in baselines, such as adversarial regularization or attention. The only trainable parameters are in the weight matrix of the 1-layer perceptron, which minimizes memory usage and improves training efficiency.


\begin{table}[tbp]
	\centering
	\caption{Experimental results of link prediction.}\label{tab:link}
	\begin{tabular}{l|cc|cc}
		\toprule
		 \multirow{2}{*}{Methods} & \multicolumn{2}{|c|}{Cora} & \multicolumn{2}{|c}{Citeseer} \\
		 \cmidrule{2-5}
		 & AUC & AP & AUC & AP\\
		 \midrule
	   GAE & 0.910 & 0.920 & 0.895 & 0.899 \\
       VGAE & 0.914 & 0.926 & 0.908 & 0.920 \\
       ARGA & 0.924 & 0.932 & 0.919 & 0.930 \\
       ARVGA & 0.924 & 0.926 & 0.924 & 0.930 \\
       GALA & 0.921 & 0.922 & 0.944 & 0.948 \\
       \midrule
       \ModelName\ & \bf 0.957 & \bf 0.952 & \bf 0.964 & \bf 0.968 \\
     \bottomrule
    \end{tabular}
\end{table}
\subsection{Link Prediction Results}

In this section, we evaluate the quality of node embeddings on the link prediction task. Following the experimental settings of GALA, we conduct experiments on Cora and Citeseer, removing 5\% edges for validation and 10\% edges for test. The training procedure and hyper-parameters remain unchanged. Given the node embedding matrix $\mathbf{Z}$, we use a simple inner product decoder to get the predicted adjacency matrix
\begin{equation}
    \hat{\mathbf{A}}=\sigma(\mathbf{ZZ}^\intercal)
\end{equation}
where $\sigma$ is the sigmoid function.

The experimental results are reported in Table~\ref{tab:link}. Compared with state-of-the-art unsupervised graph representation learning models, \ModelName\ outperforms them on both AUC and AP. It is worth noting that the training objectives of GAE/VGAE and ARGA/ARVGA are the adjacency matrix reconstruction loss. GALA also adds reconstruction loss for the link prediction task, while AGE does not utilize explicit links for supervision. 

\subsection{GAE v.s. LS+RA}
\label{sec:vs}
We use controlled experiments to verify hypothesis \textbf{H1}, evaluating the influence of entanglement of the filters and weight matrices. The compared methods are GAE and LS+RA, where the only difference between them is the position of the weight matrices. GAE, as we show in Figure~\ref{fig:gae}, combines the filter and weight matrix in each layer. LS+RA, however, moves weight matrices after the filter. Specifically, GAE has multiple GCN layers where each one contains a 64-dimensional linear layer, a ReLU activition layer and a graph convolutional filter. LS+RA stacks multiple graph convolutional filters and after which is a 1-layer 64-dimensional perceptron. Both embedding layers of the two models are 16-dimensional. Rest of the parameters are set to the same.

We report the NMI scores for node clustering on the four datasets with different number of filter layers in Figure~\ref{fig:vs}. The results show that LS+RA outperforms GAE under most circumstances with fewer parameters. Moreover, the performance of GAE decreases significantly as the filter layer increases, while LS+RA is relatively stable. A reasonable explanation to this phenomenon is stacking multiple graph convolution layers makes it harder to train all the weight matrices well. Also, the training efficiency will be affected by the deep network.
\begin{figure*}[htbp]
    \centering
    \begin{subfigure}[h]{0.24\linewidth}
    \includegraphics[width=\linewidth]{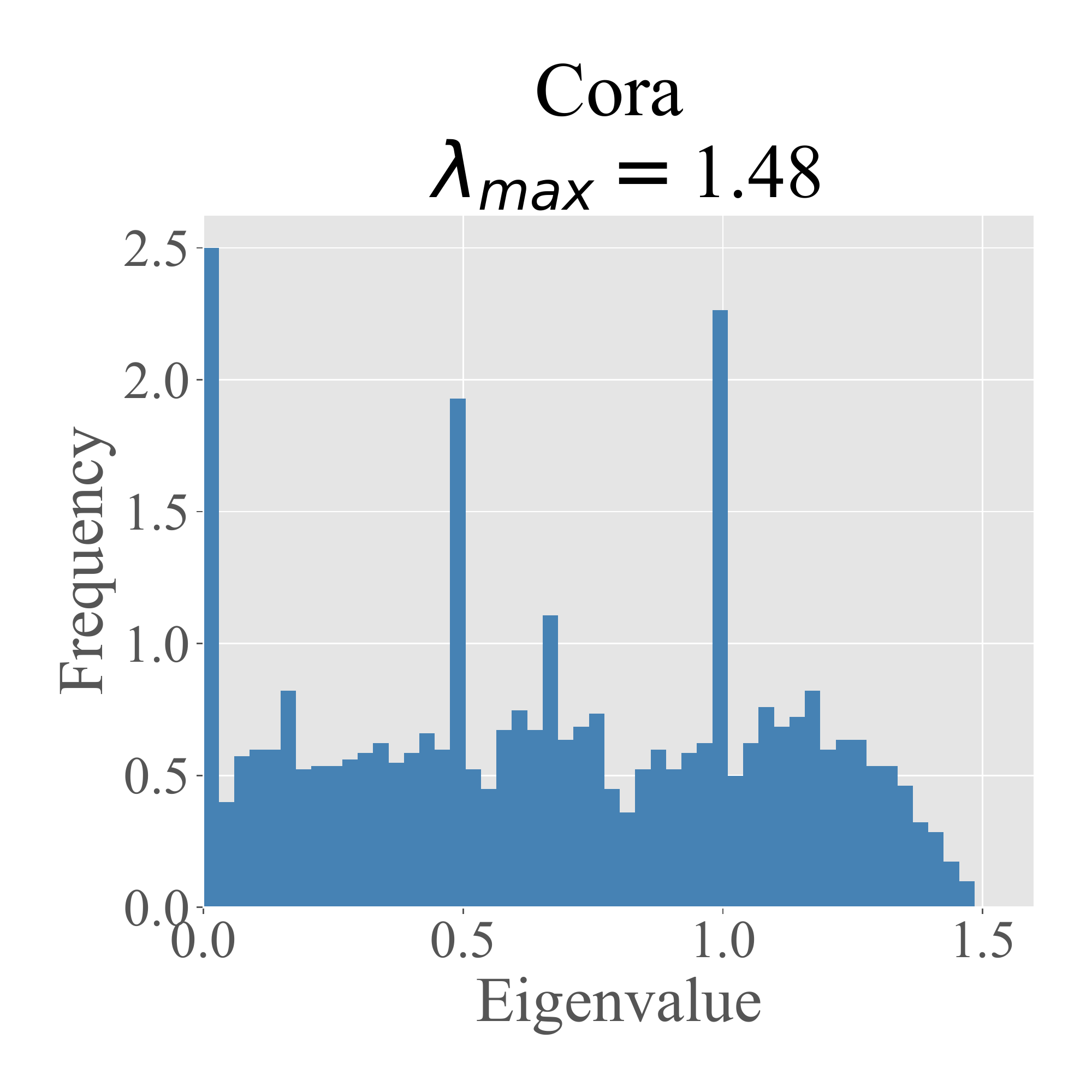}
    \end{subfigure}
    \begin{subfigure}[h]{0.24\linewidth}
    \includegraphics[width=\linewidth]{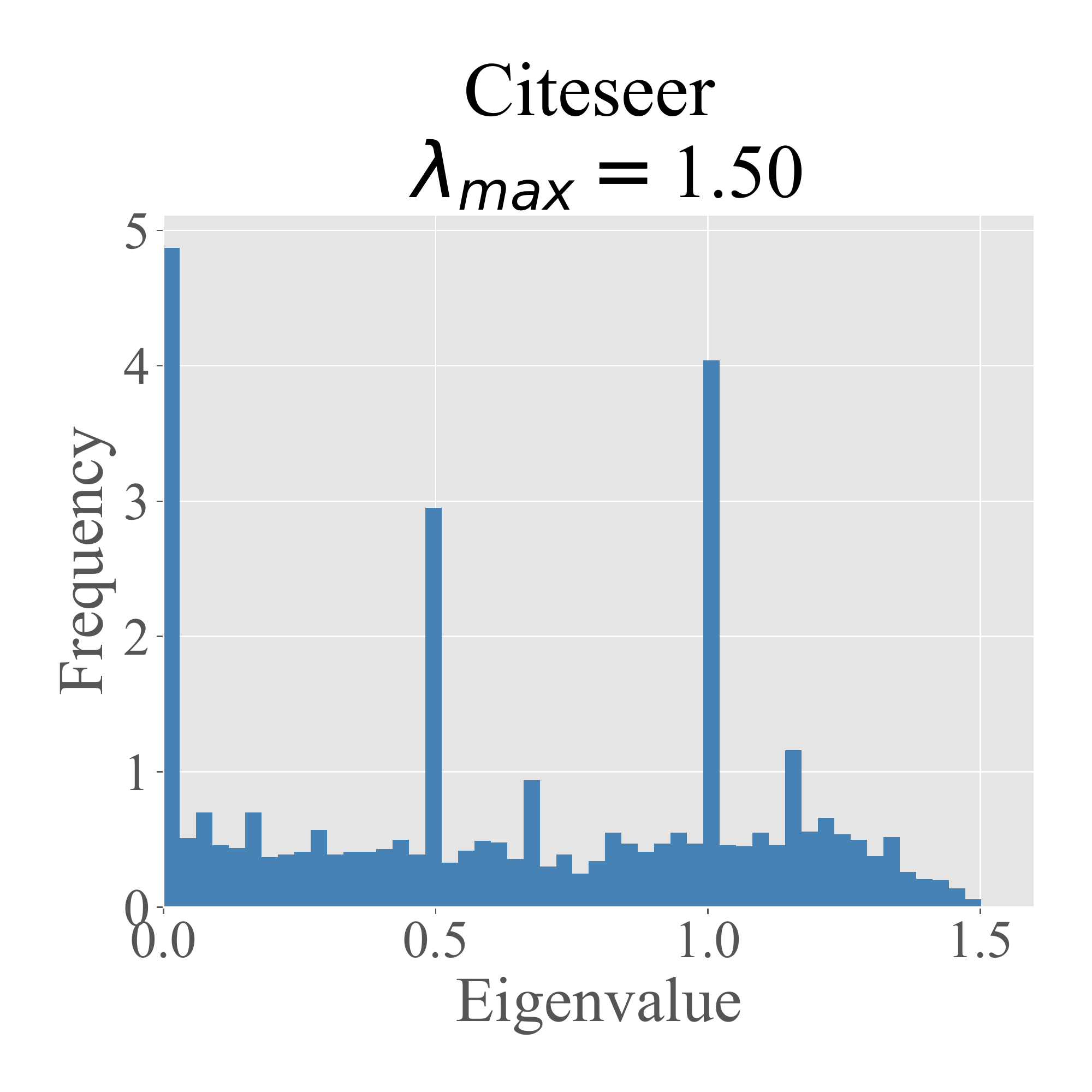}
    \end{subfigure}
    \begin{subfigure}[h]{0.24\linewidth}
    \includegraphics[width=\linewidth]{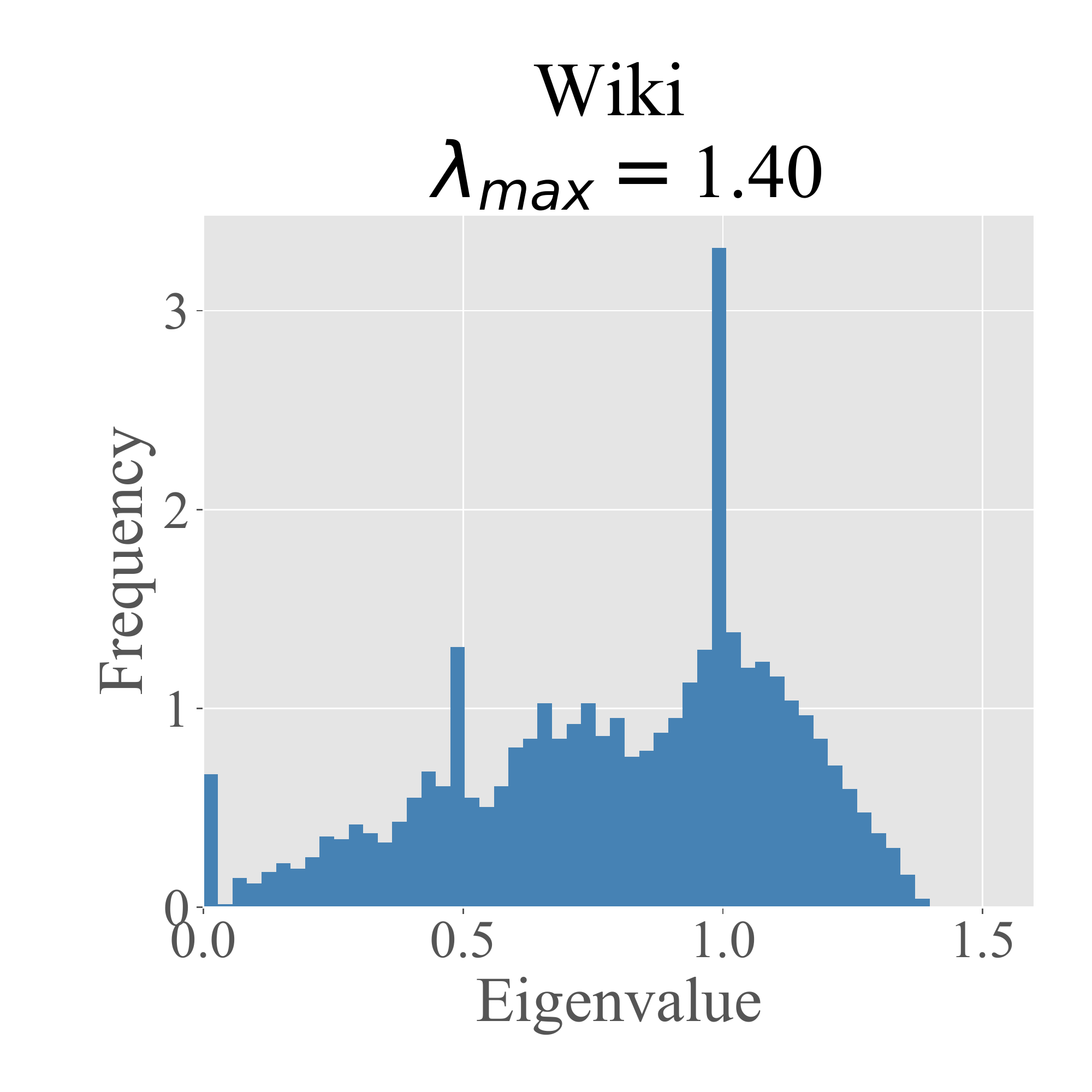}
    \end{subfigure}
    \begin{subfigure}[h]{0.24\linewidth}
    \includegraphics[width=\linewidth]{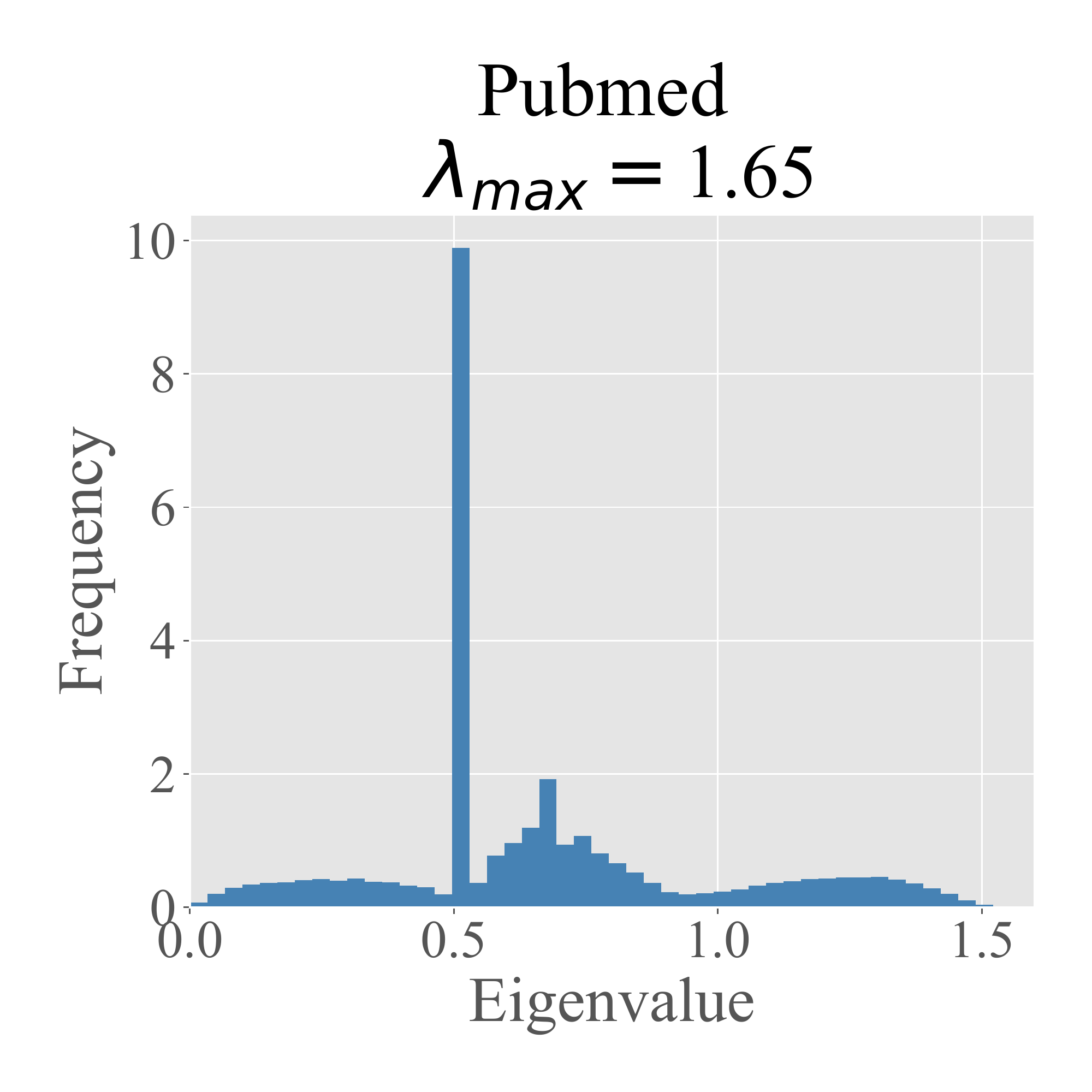}
    \end{subfigure}
    \caption{The eigenvalue distributions of benchmark datasets. $\lambda_{max}$ is the maximum eigenvalue.}
    \label{fig:eig}
\end{figure*}

\begin{figure*}[htbp]
    \centering
    \begin{subfigure}[h]{0.33\linewidth}
    \includegraphics[width=\linewidth]{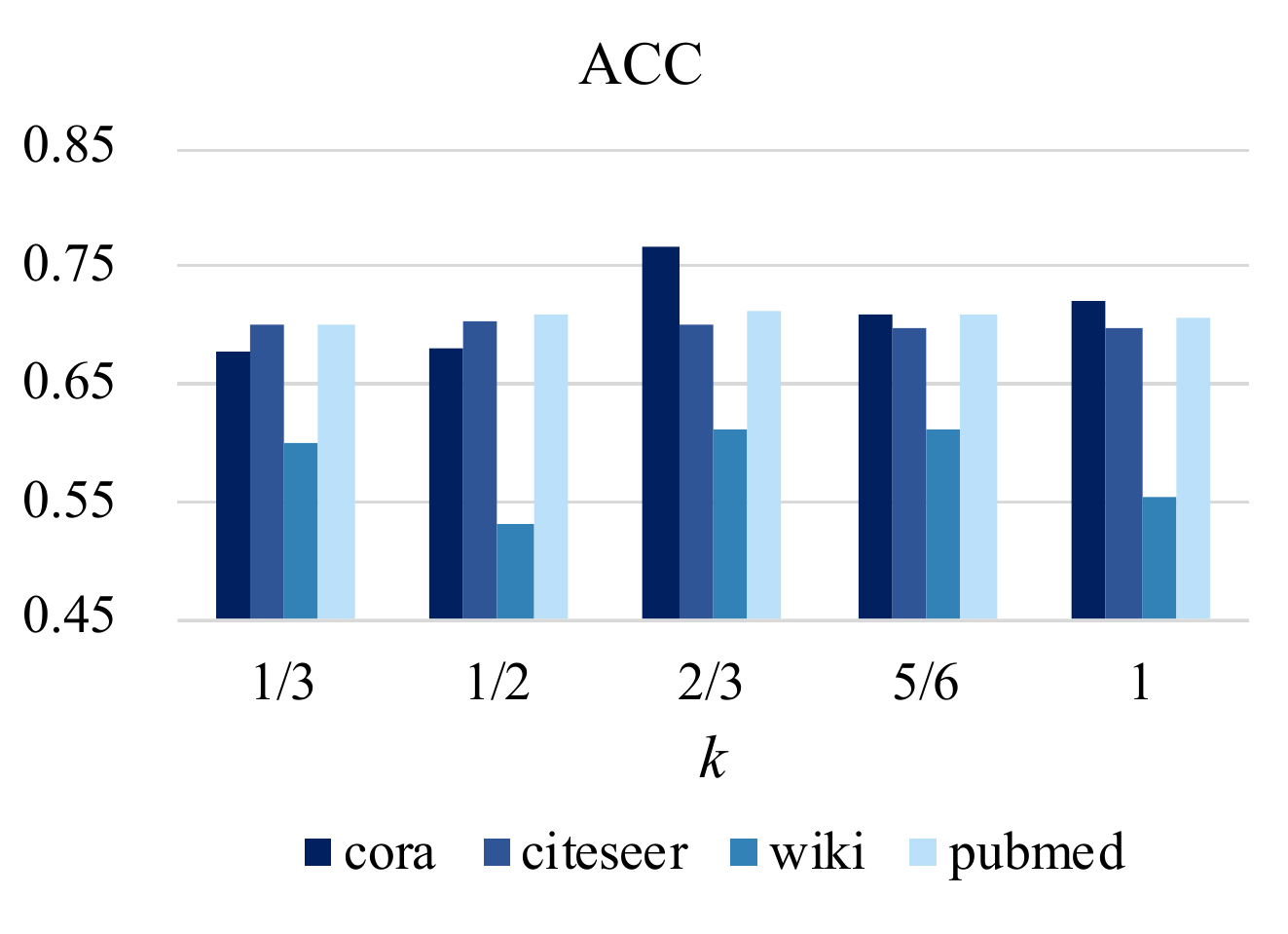}
    \end{subfigure}
    \begin{subfigure}[h]{0.33\linewidth}
    \includegraphics[width=\linewidth]{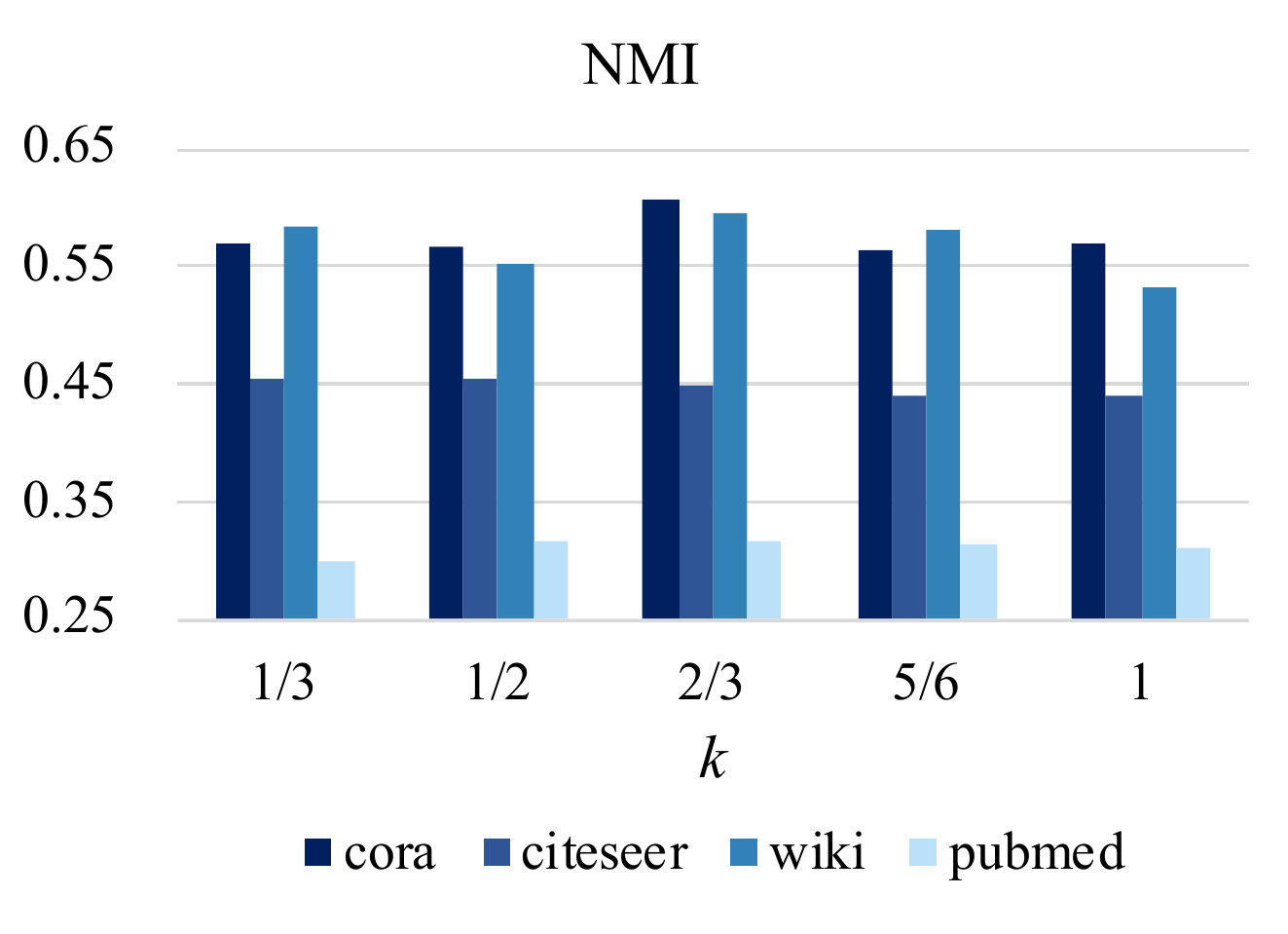}
    \end{subfigure}
    \begin{subfigure}[h]{0.33\linewidth}
    \includegraphics[width=\linewidth]{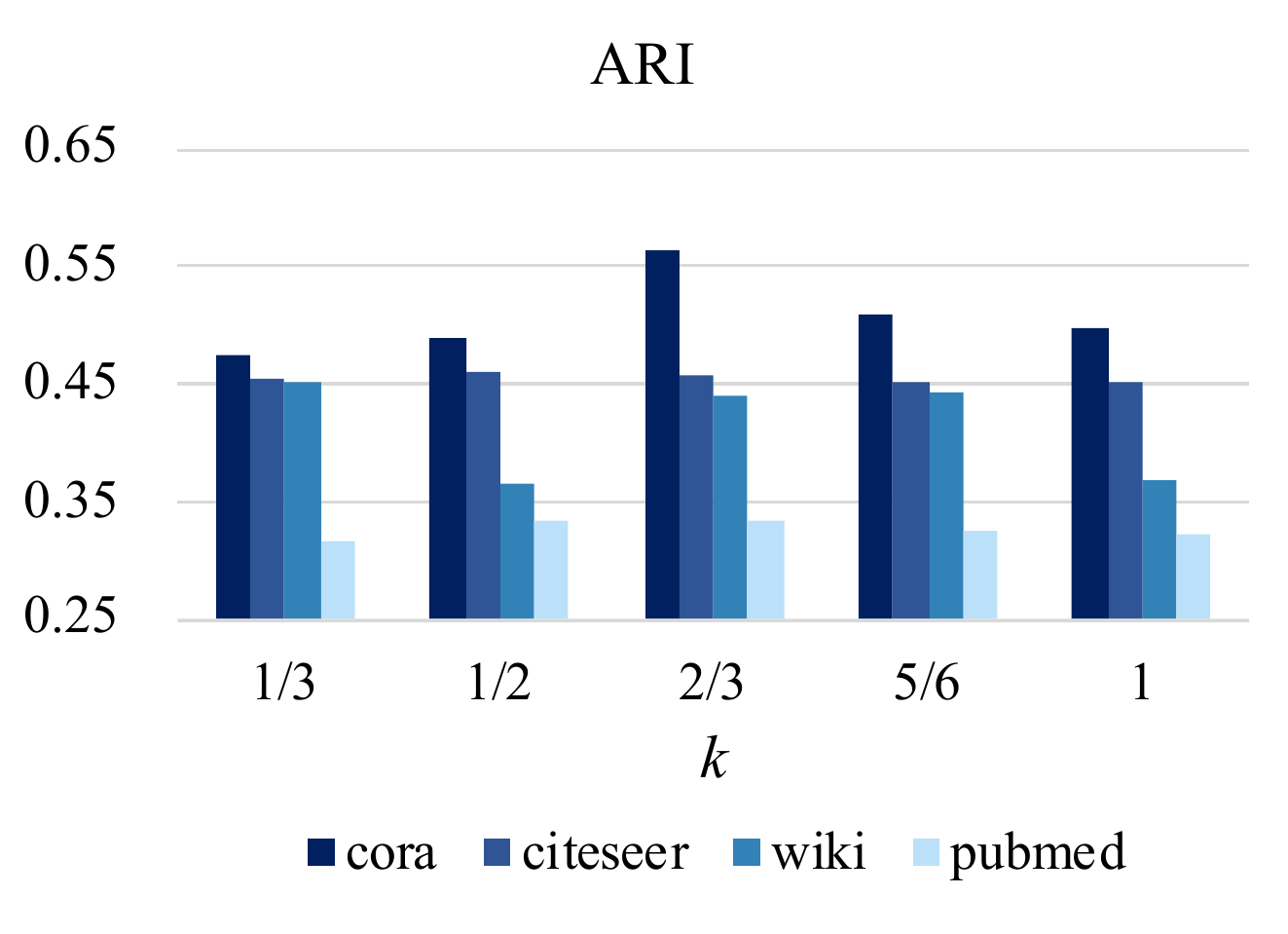}
    \end{subfigure}
    \caption{Influence of $k$ on the three metrics.}
    \label{fig:k}
\end{figure*}

\subsection{Ablation Study}
To validate \textbf{H2}, we first compare the four variants of \ModelName\ on the node clustering task. Our findings are listed below:

(1) Compared with raw features (Spectral-F), smoothed features (LS) integrate graph structure, thus perform better on node clustering. The improvement is considerable.

(2) The variants of our model, LS+RA and LS+RX, also show powerful performances compared with baseline methods, which results from our Laplacian smoothing filter. At the same time, \ModelName\ still outperforms the two variants, demonstrating that the adaptive optimization target is superior. 

(3) Comparing the two reconstruction losses, reconstructing the adjacency matrix (LS+RA) performs better on Cora, Wiki and Pubmed, while reconstructing the feature matrix (LS+RX) performs better on Citeseer. Such difference illustrates that structure information and feature information are of different importance across datasets, therefore either of them is not optimal universally. 
Furthermore, on Citeseer and Pubmed, the reconstruction losses contribute negatively to the smoothed features.

Then, we conduct ablation study on Cora to manifest the efficacy of four mechanisms in \ModelName. We set five variants of our model for comparison. 

All five variants cluster nodes by performing Spectral Clustering on the cosine similarity matrix of node features or embeddings. ``Raw features" simply performs Spectral Clustering on raw node features; ``+Filter" clusters nodes using smoothed node features; ``+Encoder" initializes training set from the similarity matrix of smoothed node features, and learns node embeddings via the fixed training set; ``+Adaptive" selects training samples adaptively with fixed thresholds; ``+Thresholds Update" further adds thresholds update strategy and is exactly the full model.

In Table~\ref{tab:abl}, it is obviously noticed that each part of our model contributes to the final performance, which evidently states the effectiveness of them. Additionally, we can observe that model supervised by the similarity of smoothed features (``+Encoder") outperforms almost all the baselines, giving verification to the rationality of our adaptive learning training objective.
\begin{table}[ht]
	\centering
	\caption{Ablation study.}\label{tab:abl}
	\begin{tabular}{l|ccc}
		\toprule
		 \multirow{2}{*}{Model Variants} & \multicolumn{3}{|c}{Cora} \\
		 \cmidrule{2-4}
		 & ACC & NMI & ARI \\
		 \midrule
       Raw features & 0.347 & 0.147 & 0.071 \\
       +Filter & 0.638 & 0.493 & 0.373 \\
       \ +Encoder & 0.728 & 0.558 & 0.521 \\
       \ \ +Adaptive & 0.739 & 0.585 & 0.544 \\
       \ \ \ +Thresholds Update & \bf 0.768 & \bf 0.607 & \bf 0.565 \\
     \bottomrule
    \end{tabular}
\end{table}

\subsection{Selection of $k$}
\label{sec:kexp}
As stated in section 3.3.3, we select $k=1/\lambda_{max}$ while $\lambda_{max}$ is the maximum eigenvalue of the renormalized Laplacian matrix. To verify the correctness of our hypothesis (\textbf{H3}), we first plot the eigenvalue distributions of the Laplacian matrix for benchmark datasets in Figure~\ref{fig:eig}. Then, we perform experiments with different $k$ and the results are report in Figure~\ref{fig:k}. From the two figures,we can make the following observations:

(1) The maximum eigenvalues of the four datasets are around $3/2$, which supports our selecting $k=2/3$.

(2) In Figure~\ref{fig:k}, it is clear that filters with $k=2/3$ work best for Cora and Wiki datasets, since all three metrics reach the highest scores at $k=2/3$. For Citeseer and Pubmed, there is little difference for various $k$.

(3) To further explain why some datasets are sensitive to $k$ while some are not, we can look back into Figure~\ref{fig:eig}. Obviously, there are more high-frequency components in Cora and Wiki than Citeseer and Pubmed. Therefore, for Citeseer and Pubmed, filters with different $k$ achieve similar effects. 

Overall, for Laplacian smoothing filters, we can conclude that $k=1/\lambda_{max}$ is the optimal choice for Laplacian smoothing filters (\textbf{H3}).

\begin{figure*}[ht]
    \centering
    \begin{subfigure}[h]{0.19\linewidth}
    \includegraphics[width=\linewidth]{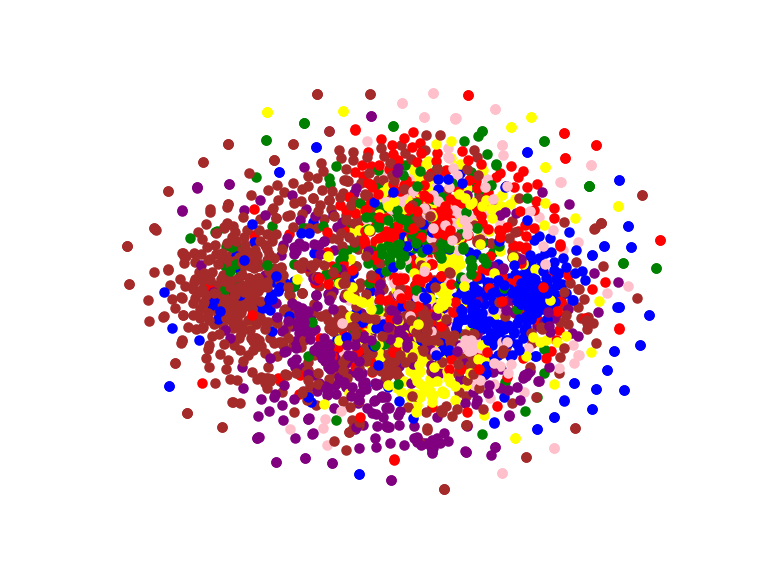}
    \caption{\label{fig:raw} Raw features}
    \end{subfigure}
    \begin{subfigure}[h]{0.19\linewidth}
    \includegraphics[width=\linewidth]{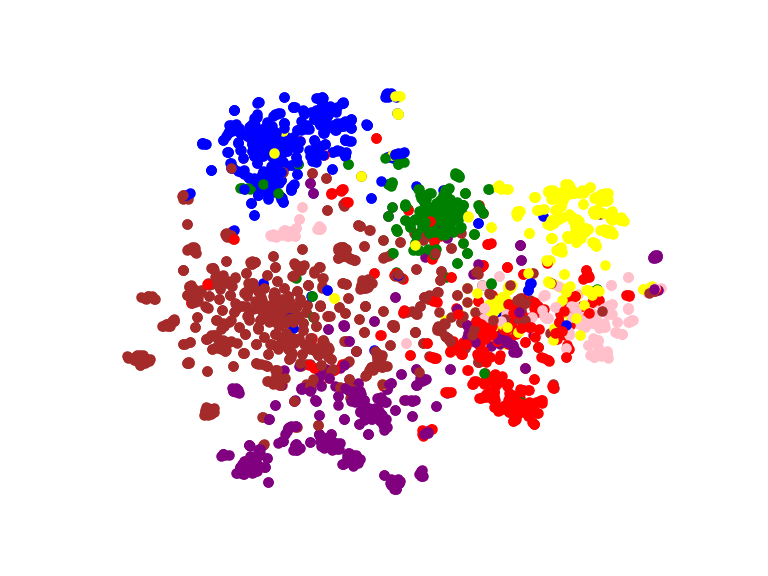}
    \caption{\label{fig:sms} +Filter}
    \end{subfigure}
    \begin{subfigure}[h]{0.19\linewidth}
    \includegraphics[width=\linewidth]{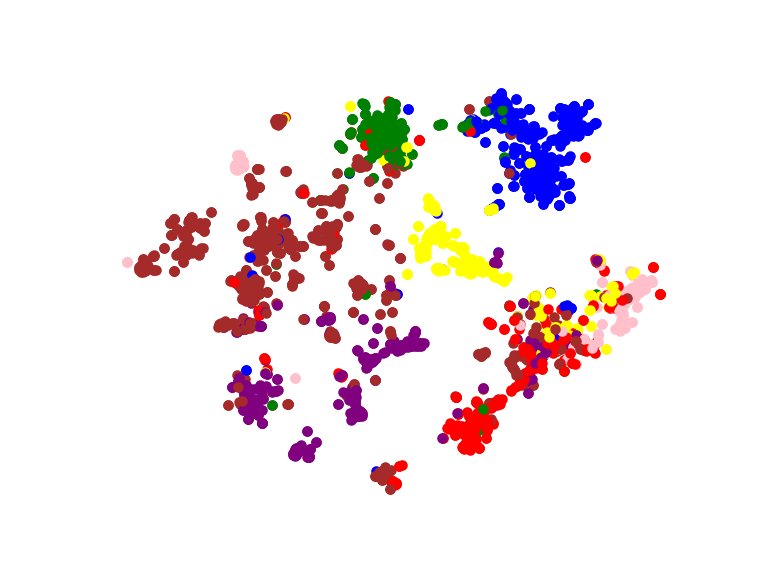}
    \caption{\label{fig:pairwise} +Encoder}
    \end{subfigure}
    \begin{subfigure}[h]{0.19\linewidth}
    \includegraphics[width=\linewidth]{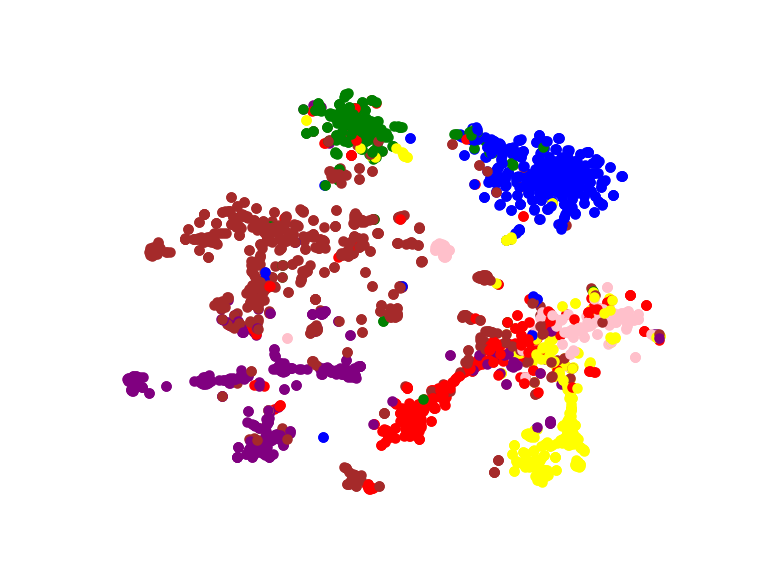}
    \caption{\label{fig:ada} +Adaptive}
    \end{subfigure}
    \begin{subfigure}[h]{0.19\linewidth}
    \includegraphics[width=\linewidth]{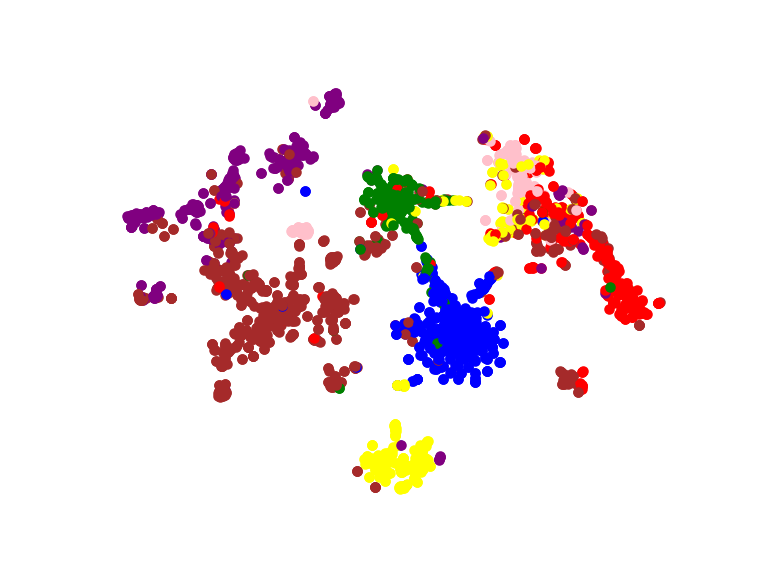}
    \caption{\label{fig:thres} Full model}
    \end{subfigure}
    \caption{2D visualization of node representations on Cora using $t$-SNE. The different colors represent different classes.}
    \label{fig:vis}
\end{figure*}

\subsection{Visualization}
To intuitively show the learned node embeddings, we visualize the node representations in 2D space using  $t$-SNE algorithm~\cite{van2014accelerating}. The figures are shown in Figure~\ref{fig:vis} and each subfigure corresponds to a variant in the ablation study. From the visualization, we can see that \ModelName\ can well cluster the nodes according to their corresponding classes. Additionally, as the model gets complete gradually, there are fewer overlapping areas and nodes belong to the same group gather together.

\section{Conclusion}
In this paper we propose \ModelName, a unified unsupervised graph representation learning model. We investigate the graph convolution operation in view of graph signal smoothing, and then design a non-parametric Laplacian smoothing filter which preserves optimal denoising properties to filter out high-frequency noises. In the encoder part, we find adaptive learning is more appropriate for embedding. Experiments on standard benchmarks demonstrate our model has outperformed state-of-the-art baseline algorithms.

For future work, an intriguing direction is to improve the computational efficiency of adaptive learning by avoiding the full computation of the pairwise similarity matrix. 

\begin{acks}
 This work is supported by the National Key Research and Development Program of China (No. 2018YFB1004503) and the National Natural Science Foundation of China (NSFC No. 61772302, 61732008).
\end{acks}
\bibliographystyle{ACM-Reference-Format}
\bibliography{KDD20AGE}


\begin{thebibliography}{39}


\ifx \showCODEN    \undefined \def \showCODEN     #1{\unskip}     \fi
\ifx \showDOI      \undefined \def \showDOI       #1{#1}\fi
\ifx \showISBNx    \undefined \def \showISBNx     #1{\unskip}     \fi
\ifx \showISBNxiii \undefined \def \showISBNxiii  #1{\unskip}     \fi
\ifx \showISSN     \undefined \def \showISSN      #1{\unskip}     \fi
\ifx \showLCCN     \undefined \def \showLCCN      #1{\unskip}     \fi
\ifx \shownote     \undefined \def \shownote      #1{#1}          \fi
\ifx \showarticletitle \undefined \def \showarticletitle #1{#1}   \fi
\ifx \showURL      \undefined \def \showURL       {\relax}        \fi
\providecommand\bibfield[2]{#2}
\providecommand\bibinfo[2]{#2}
\providecommand\natexlab[1]{#1}
\providecommand\showeprint[2][]{arXiv:#2}

\bibitem[\protect\citeauthoryear{Bengio, Louradour, Collobert, and
  Weston}{Bengio et~al\mbox{.}}{2009}]%
        {bengio2009curriculum}
\bibfield{author}{\bibinfo{person}{Yoshua Bengio},
  \bibinfo{person}{J{\'e}r{\^o}me Louradour}, \bibinfo{person}{Ronan
  Collobert}, {and} \bibinfo{person}{Jason Weston}.}
  \bibinfo{year}{2009}\natexlab{}.
\newblock \showarticletitle{Curriculum learning}. In
  \bibinfo{booktitle}{\emph{Proceedings of ICML}}. \bibinfo{pages}{41--48}.
\newblock


\bibitem[\protect\citeauthoryear{Bojchevski and G{\"u}nnemann}{Bojchevski and
  G{\"u}nnemann}{2018}]%
        {bojchevski2018bayesian}
\bibfield{author}{\bibinfo{person}{Aleksandar Bojchevski} {and}
  \bibinfo{person}{Stephan G{\"u}nnemann}.} \bibinfo{year}{2018}\natexlab{}.
\newblock \showarticletitle{Bayesian robust attributed graph clustering: Joint
  learning of partial anomalies and group structure}. In
  \bibinfo{booktitle}{\emph{Proceedings of AAAI}}. \bibinfo{pages}{2738--2745}.
\newblock


\bibitem[\protect\citeauthoryear{Cao, Lu, and Xu}{Cao et~al\mbox{.}}{2015}]%
        {cao2015grarep}
\bibfield{author}{\bibinfo{person}{Shaosheng Cao}, \bibinfo{person}{Wei Lu},
  {and} \bibinfo{person}{Qiongkai Xu}.} \bibinfo{year}{2015}\natexlab{}.
\newblock \showarticletitle{Grarep: Learning graph representations with global
  structural information}. In \bibinfo{booktitle}{\emph{Proceedings of CIKM}}.
  \bibinfo{pages}{891--900}.
\newblock


\bibitem[\protect\citeauthoryear{Cao, Lu, and Xu}{Cao et~al\mbox{.}}{2016}]%
        {cao2016deep}
\bibfield{author}{\bibinfo{person}{Shaosheng Cao}, \bibinfo{person}{Wei Lu},
  {and} \bibinfo{person}{Qiongkai Xu}.} \bibinfo{year}{2016}\natexlab{}.
\newblock \showarticletitle{Deep neural networks for learning graph
  representations}. In \bibinfo{booktitle}{\emph{Proceedings of AAAI}}.
  \bibinfo{pages}{1145--1152}.
\newblock


\bibitem[\protect\citeauthoryear{Chang and Blei}{Chang and Blei}{2009}]%
        {chang2009relational}
\bibfield{author}{\bibinfo{person}{Jonathan Chang} {and} \bibinfo{person}{David
  Blei}.} \bibinfo{year}{2009}\natexlab{}.
\newblock \showarticletitle{Relational topic models for document networks}. In
  \bibinfo{booktitle}{\emph{Artificial Intelligence and Statistics}}.
  \bibinfo{pages}{81--88}.
\newblock


\bibitem[\protect\citeauthoryear{Chang, Wang, Meng, Xiang, and Pan}{Chang
  et~al\mbox{.}}{2017}]%
        {chang2017deep}
\bibfield{author}{\bibinfo{person}{Jianlong Chang}, \bibinfo{person}{Lingfeng
  Wang}, \bibinfo{person}{Gaofeng Meng}, \bibinfo{person}{Shiming Xiang}, {and}
  \bibinfo{person}{Chunhong Pan}.} \bibinfo{year}{2017}\natexlab{}.
\newblock \showarticletitle{Deep adaptive image clustering}. In
  \bibinfo{booktitle}{\emph{Proceedings of ICCV}}. \bibinfo{pages}{5879--5887}.
\newblock


\bibitem[\protect\citeauthoryear{Chung and Graham}{Chung and Graham}{1997}]%
        {chung1997spectral}
\bibfield{author}{\bibinfo{person}{Fan~RK Chung} {and}
  \bibinfo{person}{Fan~Chung Graham}.} \bibinfo{year}{1997}\natexlab{}.
\newblock \bibinfo{booktitle}{\emph{Spectral graph theory}}.
\newblock \bibinfo{publisher}{American Mathematical Soc.}
\newblock


\bibitem[\protect\citeauthoryear{Davies and Bouldin}{Davies and
  Bouldin}{1979}]%
        {davies1979cluster}
\bibfield{author}{\bibinfo{person}{David~L Davies} {and}
  \bibinfo{person}{Donald~W Bouldin}.} \bibinfo{year}{1979}\natexlab{}.
\newblock \showarticletitle{A cluster separation measure}.
\newblock \bibinfo{journal}{\emph{IEEE transactions on pattern analysis and
  machine intelligence}} \bibinfo{number}{2} (\bibinfo{year}{1979}),
  \bibinfo{pages}{224--227}.
\newblock


\bibitem[\protect\citeauthoryear{Gan, Ma, and Wu}{Gan et~al\mbox{.}}{2007}]%
        {gan2007data}
\bibfield{author}{\bibinfo{person}{Guojun Gan}, \bibinfo{person}{Chaoqun Ma},
  {and} \bibinfo{person}{Jianhong Wu}.} \bibinfo{year}{2007}\natexlab{}.
\newblock \bibinfo{booktitle}{\emph{Data clustering: Theory, algorithms, and
  applications}}. Vol.~\bibinfo{volume}{20}.
\newblock \bibinfo{publisher}{Siam}.
\newblock


\bibitem[\protect\citeauthoryear{Grover and Leskovec}{Grover and
  Leskovec}{2016}]%
        {grover2016node2vec}
\bibfield{author}{\bibinfo{person}{Aditya Grover} {and} \bibinfo{person}{Jure
  Leskovec}.} \bibinfo{year}{2016}\natexlab{}.
\newblock \showarticletitle{node2vec: Scalable feature learning for networks}.
  In \bibinfo{booktitle}{\emph{Proceedings of SIGKDD}}.
  \bibinfo{pages}{855--864}.
\newblock


\bibitem[\protect\citeauthoryear{Hamilton, Ying, and Leskovec}{Hamilton
  et~al\mbox{.}}{2017}]%
        {hamilton2017representation}
\bibfield{author}{\bibinfo{person}{William~L. Hamilton}, \bibinfo{person}{Rex
  Ying}, {and} \bibinfo{person}{Jure Leskovec}.}
  \bibinfo{year}{2017}\natexlab{}.
\newblock \showarticletitle{Representation learning on graphs: Methods and
  applications}.
\newblock \bibinfo{journal}{\emph{IEEE Data(base) Engineering Bulletin}}
  \bibinfo{volume}{40}, \bibinfo{number}{3} (\bibinfo{year}{2017}),
  \bibinfo{pages}{52--74}.
\newblock


\bibitem[\protect\citeauthoryear{Hastings}{Hastings}{2006}]%
        {hastings2006community}
\bibfield{author}{\bibinfo{person}{Matthew~B Hastings}.}
  \bibinfo{year}{2006}\natexlab{}.
\newblock \showarticletitle{Community detection as an inference problem}.
\newblock \bibinfo{journal}{\emph{Physical Review E}} \bibinfo{volume}{74},
  \bibinfo{number}{3} (\bibinfo{year}{2006}), \bibinfo{pages}{035--102}.
\newblock


\bibitem[\protect\citeauthoryear{Horn and Johnson}{Horn and Johnson}{2012}]%
        {horn2012matrix}
\bibfield{author}{\bibinfo{person}{Roger~A Horn} {and}
  \bibinfo{person}{Charles~R Johnson}.} \bibinfo{year}{2012}\natexlab{}.
\newblock \bibinfo{booktitle}{\emph{Matrix analysis}}.
\newblock \bibinfo{publisher}{Cambridge university press}.
\newblock


\bibitem[\protect\citeauthoryear{Kingma and Ba}{Kingma and Ba}{2015}]%
        {kingma2014adam}
\bibfield{author}{\bibinfo{person}{Diederik~P Kingma} {and}
  \bibinfo{person}{Jimmy Ba}.} \bibinfo{year}{2015}\natexlab{}.
\newblock \showarticletitle{Adam: A method for stochastic optimization}. In
  \bibinfo{booktitle}{\emph{Proceedings of ICLR}}. 15.
\newblock


\bibitem[\protect\citeauthoryear{Kipf and Welling}{Kipf and Welling}{2016}]%
        {kipf2016variational}
\bibfield{author}{\bibinfo{person}{Thomas~N Kipf} {and} \bibinfo{person}{Max
  Welling}.} \bibinfo{year}{2016}\natexlab{}.
\newblock \showarticletitle{Variational graph auto-encoders}. In
  \bibinfo{booktitle}{\emph{NIPS Workshop on Bayesian Deep Learning}}. 3.
\newblock


\bibitem[\protect\citeauthoryear{Kipf and Welling}{Kipf and Welling}{2017}]%
        {kipf2016semi}
\bibfield{author}{\bibinfo{person}{Thomas~N Kipf} {and} \bibinfo{person}{Max
  Welling}.} \bibinfo{year}{2017}\natexlab{}.
\newblock \showarticletitle{Semi-supervised classification with graph
  convolutional networks}. In \bibinfo{booktitle}{\emph{Proceedings of ICLR}}.
  14.
\newblock


\bibitem[\protect\citeauthoryear{Le and Mikolov}{Le and Mikolov}{2014}]%
        {le2014distributed}
\bibfield{author}{\bibinfo{person}{Quoc Le} {and} \bibinfo{person}{Tomas
  Mikolov}.} \bibinfo{year}{2014}\natexlab{}.
\newblock \showarticletitle{Distributed representations of sentences and
  documents}. In \bibinfo{booktitle}{\emph{Proceedings of ICML}}.
  \bibinfo{pages}{1188--1196}.
\newblock


\bibitem[\protect\citeauthoryear{Li, Han, and Wu}{Li et~al\mbox{.}}{2018a}]%
        {li2018deeper}
\bibfield{author}{\bibinfo{person}{Qimai Li}, \bibinfo{person}{Zhichao Han},
  {and} \bibinfo{person}{Xiao-Ming Wu}.} \bibinfo{year}{2018}\natexlab{a}.
\newblock \showarticletitle{Deeper insights into graph convolutional networks
  for semi-supervised learning}. In \bibinfo{booktitle}{\emph{Proceedings of
  AAAI}}. \bibinfo{pages}{3538--3545}.
\newblock


\bibitem[\protect\citeauthoryear{Li, Sha, Huang, and Zhang}{Li
  et~al\mbox{.}}{2018b}]%
        {li2018community}
\bibfield{author}{\bibinfo{person}{Ye Li}, \bibinfo{person}{Chaofeng Sha},
  \bibinfo{person}{Xin Huang}, {and} \bibinfo{person}{Yanchun Zhang}.}
  \bibinfo{year}{2018}\natexlab{b}.
\newblock \showarticletitle{Community detection in attributed graphs: an
  embedding approach}. In \bibinfo{booktitle}{\emph{Proceedings of AAAI}}.
  \bibinfo{pages}{338--345}.
\newblock


\bibitem[\protect\citeauthoryear{Lloyd}{Lloyd}{1982}]%
        {lloyd1982least}
\bibfield{author}{\bibinfo{person}{Stuart Lloyd}.}
  \bibinfo{year}{1982}\natexlab{}.
\newblock \showarticletitle{Least squares quantization in PCM}.
\newblock \bibinfo{journal}{\emph{IEEE transactions on information theory}}
  \bibinfo{volume}{28}, \bibinfo{number}{2} (\bibinfo{year}{1982}),
  \bibinfo{pages}{129--137}.
\newblock


\bibitem[\protect\citeauthoryear{Newman}{Newman}{2006}]%
        {newman2006finding}
\bibfield{author}{\bibinfo{person}{Mark~EJ Newman}.}
  \bibinfo{year}{2006}\natexlab{}.
\newblock \showarticletitle{Finding community structure in networks using the
  eigenvectors of matrices}.
\newblock \bibinfo{journal}{\emph{Physical review E}} \bibinfo{volume}{74},
  \bibinfo{number}{3} (\bibinfo{year}{2006}), \bibinfo{pages}{036--104}.
\newblock


\bibitem[\protect\citeauthoryear{Ng, Jordan, and Weiss}{Ng
  et~al\mbox{.}}{2002}]%
        {ng2002spectral}
\bibfield{author}{\bibinfo{person}{Andrew~Y Ng}, \bibinfo{person}{Michael~I
  Jordan}, {and} \bibinfo{person}{Yair Weiss}.}
  \bibinfo{year}{2002}\natexlab{}.
\newblock \showarticletitle{On spectral clustering: Analysis and an algorithm}.
  In \bibinfo{booktitle}{\emph{Proceedings of NIPS}}.
  \bibinfo{pages}{849--856}.
\newblock


\bibitem[\protect\citeauthoryear{Pan, Hu, Long, Jiang, Yao, and Zhang}{Pan
  et~al\mbox{.}}{2018}]%
        {pan2018adversarially}
\bibfield{author}{\bibinfo{person}{Shirui Pan}, \bibinfo{person}{Ruiqi Hu},
  \bibinfo{person}{Guodong Long}, \bibinfo{person}{Jing Jiang},
  \bibinfo{person}{Lina Yao}, {and} \bibinfo{person}{Chengqi Zhang}.}
  \bibinfo{year}{2018}\natexlab{}.
\newblock \showarticletitle{Adversarially regularized graph autoencoder for
  graph embedding}. In \bibinfo{booktitle}{\emph{Proceedings of IJCAI}}.
  \bibinfo{pages}{2609--2615}.
\newblock


\bibitem[\protect\citeauthoryear{Park, Lee, Chang, Lee, and Choi}{Park
  et~al\mbox{.}}{2019}]%
        {park2019symmetric}
\bibfield{author}{\bibinfo{person}{Jiwoong Park}, \bibinfo{person}{Minsik Lee},
  \bibinfo{person}{Hyung~Jin Chang}, \bibinfo{person}{Kyuewang Lee}, {and}
  \bibinfo{person}{Jin~Young Choi}.} \bibinfo{year}{2019}\natexlab{}.
\newblock \showarticletitle{Symmetric graph convolutional autoencoder for
  unsupervised graph representation learning}. In
  \bibinfo{booktitle}{\emph{Proceedings of ICCV}}. \bibinfo{pages}{6519--6528}.
\newblock


\bibitem[\protect\citeauthoryear{Perozzi, Al-Rfou, and Skiena}{Perozzi
  et~al\mbox{.}}{2014}]%
        {perozzi2014deepwalk}
\bibfield{author}{\bibinfo{person}{Bryan Perozzi}, \bibinfo{person}{Rami
  Al-Rfou}, {and} \bibinfo{person}{Steven Skiena}.}
  \bibinfo{year}{2014}\natexlab{}.
\newblock \showarticletitle{Deepwalk: Online learning of social
  representations}. In \bibinfo{booktitle}{\emph{Proceedings of SIGKDD}}.
  \bibinfo{pages}{701--710}.
\newblock


\bibitem[\protect\citeauthoryear{Sen, Namata, Bilgic, Getoor, Galligher, and
  Eliassi-Rad}{Sen et~al\mbox{.}}{2008}]%
        {sen2008collective}
\bibfield{author}{\bibinfo{person}{Prithviraj Sen}, \bibinfo{person}{Galileo
  Namata}, \bibinfo{person}{Mustafa Bilgic}, \bibinfo{person}{Lise Getoor},
  \bibinfo{person}{Brian Galligher}, {and} \bibinfo{person}{Tina Eliassi-Rad}.}
  \bibinfo{year}{2008}\natexlab{}.
\newblock \showarticletitle{Collective classification in network data}.
\newblock \bibinfo{journal}{\emph{AI magazine}} \bibinfo{volume}{29},
  \bibinfo{number}{3} (\bibinfo{year}{2008}), \bibinfo{pages}{93--93}.
\newblock


\bibitem[\protect\citeauthoryear{Tang, Qu, Wang, Zhang, Yan, and Mei}{Tang
  et~al\mbox{.}}{2015}]%
        {tang2015line}
\bibfield{author}{\bibinfo{person}{Jian Tang}, \bibinfo{person}{Meng Qu},
  \bibinfo{person}{Mingzhe Wang}, \bibinfo{person}{Ming Zhang},
  \bibinfo{person}{Jun Yan}, {and} \bibinfo{person}{Qiaozhu Mei}.}
  \bibinfo{year}{2015}\natexlab{}.
\newblock \showarticletitle{Line: Large-scale information network embedding}.
  In \bibinfo{booktitle}{\emph{Proceedings of WWW}}.
  \bibinfo{pages}{1067--1077}.
\newblock


\bibitem[\protect\citeauthoryear{Taubin}{Taubin}{1995}]%
        {taubin1995signal}
\bibfield{author}{\bibinfo{person}{Gabriel Taubin}.}
  \bibinfo{year}{1995}\natexlab{}.
\newblock \showarticletitle{A signal processing approach to fair surface
  design}. In \bibinfo{booktitle}{\emph{Proceedings of the 22nd annual
  conference on Computer graphics and interactive techniques}}.
  \bibinfo{pages}{351--358}.
\newblock


\bibitem[\protect\citeauthoryear{Van Der~Maaten}{Van Der~Maaten}{2014}]%
        {van2014accelerating}
\bibfield{author}{\bibinfo{person}{Laurens Van Der~Maaten}.}
  \bibinfo{year}{2014}\natexlab{}.
\newblock \showarticletitle{Accelerating t-SNE using tree-based algorithms}.
\newblock \bibinfo{journal}{\emph{The journal of machine learning research}}
  \bibinfo{volume}{15}, \bibinfo{number}{1} (\bibinfo{year}{2014}),
  \bibinfo{pages}{3221--3245}.
\newblock


\bibitem[\protect\citeauthoryear{Veli{\v{c}}kovi{\'c}, Cucurull, Casanova,
  Romero, Lio, and Bengio}{Veli{\v{c}}kovi{\'c} et~al\mbox{.}}{2018}]%
        {velivckovic2017graph}
\bibfield{author}{\bibinfo{person}{Petar Veli{\v{c}}kovi{\'c}},
  \bibinfo{person}{Guillem Cucurull}, \bibinfo{person}{Arantxa Casanova},
  \bibinfo{person}{Adriana Romero}, \bibinfo{person}{Pietro Lio}, {and}
  \bibinfo{person}{Yoshua Bengio}.} \bibinfo{year}{2018}\natexlab{}.
\newblock \showarticletitle{Graph attention networks}. In
  \bibinfo{booktitle}{\emph{Proceedings of ICLR}}. 8.
\newblock


\bibitem[\protect\citeauthoryear{Wang, Pan, Hu, Long, Jiang, and Zhang}{Wang
  et~al\mbox{.}}{2019}]%
        {wang2019attributed}
\bibfield{author}{\bibinfo{person}{Chun Wang}, \bibinfo{person}{Shirui Pan},
  \bibinfo{person}{Ruiqi Hu}, \bibinfo{person}{Guodong Long},
  \bibinfo{person}{Jing Jiang}, {and} \bibinfo{person}{Chengqi Zhang}.}
  \bibinfo{year}{2019}\natexlab{}.
\newblock \showarticletitle{Attributed graph clustering: A deep attentional
  embedding approach}. In \bibinfo{booktitle}{\emph{Proceedings of IJCAI}}.
  \bibinfo{pages}{3670--3676}.
\newblock


\bibitem[\protect\citeauthoryear{Wang, Pan, Long, Zhu, and Jiang}{Wang
  et~al\mbox{.}}{2017}]%
        {wang2017mgae}
\bibfield{author}{\bibinfo{person}{Chun Wang}, \bibinfo{person}{Shirui Pan},
  \bibinfo{person}{Guodong Long}, \bibinfo{person}{Xingquan Zhu}, {and}
  \bibinfo{person}{Jing Jiang}.} \bibinfo{year}{2017}\natexlab{}.
\newblock \showarticletitle{Mgae: Marginalized graph autoencoder for graph
  clustering}. In \bibinfo{booktitle}{\emph{Proceedings of CIKM}}.
  \bibinfo{pages}{889--898}.
\newblock


\bibitem[\protect\citeauthoryear{Wang, Cui, and Zhu}{Wang
  et~al\mbox{.}}{2016a}]%
        {wang2016structural}
\bibfield{author}{\bibinfo{person}{Daixin Wang}, \bibinfo{person}{Peng Cui},
  {and} \bibinfo{person}{Wenwu Zhu}.} \bibinfo{year}{2016}\natexlab{a}.
\newblock \showarticletitle{Structural deep network embedding}. In
  \bibinfo{booktitle}{\emph{Proceedings of SIGKDD}}.
  \bibinfo{pages}{1225--1234}.
\newblock


\bibitem[\protect\citeauthoryear{Wang, Jin, Cao, Yang, and Zhang}{Wang
  et~al\mbox{.}}{2016b}]%
        {wang2016semantic}
\bibfield{author}{\bibinfo{person}{Xiao Wang}, \bibinfo{person}{Di Jin},
  \bibinfo{person}{Xiaochun Cao}, \bibinfo{person}{Liang Yang}, {and}
  \bibinfo{person}{Weixiong Zhang}.} \bibinfo{year}{2016}\natexlab{b}.
\newblock \showarticletitle{Semantic community identification in large
  attribute networks}. In \bibinfo{booktitle}{\emph{Proceedings of AAAI}}.
  \bibinfo{pages}{265–--271}.
\newblock


\bibitem[\protect\citeauthoryear{Wu, Zhang, Souza~Jr, Fifty, Yu, and
  Weinberger}{Wu et~al\mbox{.}}{2019}]%
        {wu2019simplifying}
\bibfield{author}{\bibinfo{person}{Felix Wu}, \bibinfo{person}{Tianyi Zhang},
  \bibinfo{person}{Amauri Holanda~de Souza~Jr}, \bibinfo{person}{Christopher
  Fifty}, \bibinfo{person}{Tao Yu}, {and} \bibinfo{person}{Kilian~Q
  Weinberger}.} \bibinfo{year}{2019}\natexlab{}.
\newblock \showarticletitle{Simplifying graph convolutional networks}. In
  \bibinfo{booktitle}{\emph{Proceedings of ICML}}. \bibinfo{pages}{6861--6871}.
\newblock


\bibitem[\protect\citeauthoryear{Yang, Liu, Zhao, Sun, and Chang}{Yang
  et~al\mbox{.}}{2015}]%
        {yang2015network}
\bibfield{author}{\bibinfo{person}{Cheng Yang}, \bibinfo{person}{Zhiyuan Liu},
  \bibinfo{person}{Deli Zhao}, \bibinfo{person}{Maosong Sun}, {and}
  \bibinfo{person}{Edward Chang}.} \bibinfo{year}{2015}\natexlab{}.
\newblock \showarticletitle{Network representation learning with rich text
  information}. In \bibinfo{booktitle}{\emph{Proceedings of IJCAI}}.
  \bibinfo{pages}{2111--2117}.
\newblock


\bibitem[\protect\citeauthoryear{Ying, He, Chen, Eksombatchai, Hamilton, and
  Leskovec}{Ying et~al\mbox{.}}{2018}]%
        {ying2018graph}
\bibfield{author}{\bibinfo{person}{Rex Ying}, \bibinfo{person}{Ruining He},
  \bibinfo{person}{Kaifeng Chen}, \bibinfo{person}{Pong Eksombatchai},
  \bibinfo{person}{William~L Hamilton}, {and} \bibinfo{person}{Jure Leskovec}.}
  \bibinfo{year}{2018}\natexlab{}.
\newblock \showarticletitle{Graph convolutional neural networks for web-scale
  recommender systems}. In \bibinfo{booktitle}{\emph{Proceedings of SIGKDD}}.
  \bibinfo{pages}{974--983}.
\newblock


\bibitem[\protect\citeauthoryear{Zhang, Liu, Li, and Wu}{Zhang
  et~al\mbox{.}}{2019}]%
        {zhang2019attributed}
\bibfield{author}{\bibinfo{person}{Xiaotong Zhang}, \bibinfo{person}{Han Liu},
  \bibinfo{person}{Qimai Li}, {and} \bibinfo{person}{Xiao-Ming Wu}.}
  \bibinfo{year}{2019}\natexlab{}.
\newblock \showarticletitle{Attributed graph clustering via adaptive graph
  convolution}. In \bibinfo{booktitle}{\emph{Proceedings of IJCAI}}.
  \bibinfo{pages}{4327--4333}.
\newblock


\bibitem[\protect\citeauthoryear{Zhou, Cui, Zhang, Yang, Liu, Wang, Li, and
  Sun}{Zhou et~al\mbox{.}}{2018}]%
        {zhou2018graph}
\bibfield{author}{\bibinfo{person}{Jie Zhou}, \bibinfo{person}{Ganqu Cui},
  \bibinfo{person}{Zhengyan Zhang}, \bibinfo{person}{Cheng Yang},
  \bibinfo{person}{Zhiyuan Liu}, \bibinfo{person}{Lifeng Wang},
  \bibinfo{person}{Changcheng Li}, {and} \bibinfo{person}{Maosong Sun}.}
  \bibinfo{year}{2018}\natexlab{}.
\newblock \showarticletitle{Graph neural networks: A review of methods and
  applications}.
\newblock \bibinfo{journal}{\emph{arXiv preprint arXiv:1812.08434}}
  (\bibinfo{year}{2018}).
\newblock


\end{thebibliography}

\appendix
\section{More Details About The Experiments}
Here we describe more details about the experiments to help in reproducibility.

\subsection{Hardware and Software Configurations}
All experiments are conducted on a server under the same environment.

Hardware:
\begin{itemize}
    \item Operating System: Ubuntu 18.04.3 LTS
    \item CPU: Intel(R) Xeon(R) Gold 5218 CPU @ 2.30GHz
    \item GPU: GeForce RTX 2080 Ti
\end{itemize}

Software:
\begin{itemize}
    \item Python 3.7.4
    \item PyTorch 1.3.1
    \item sklearn 0.21.3
\end{itemize}
\subsection{Hyperparameter Settings}
We report our hyperparameter settings in Table~\ref{tab:param}.
\\

\captionof{table}{\textbf{Hyperparameter settings, where $n$ is number of nodes in the dataset.}}
    \begin{tabular}{l|ccccc}
    \toprule
        Dataset & $t$ & $r_{pos}^{st}/n^2$ & $r_{pos}^{ed}/n^2$ & $r_{neg}^{st}/n^2$ & $r_{neg}^{ed}/n^2$ \\
    \midrule
       Cora & 8 & 0.0110 & 0.0010 & 0.1 & 0.5 \\
       Citeseer & 3 & 0.0015 & 0.0010 & 0.1 & 0.5\\
       Wiki & 1 & 0.0011 & 0.0010 & 0.1 & 0.5\\
       Pubmed & 35 & 0.0013 & 0.0010 & 0.7 & 0.8\\
     \bottomrule
    \end{tabular}
    \label{tab:param}
\end{document}